%% file: main.tex
\documentclass{article}

\usepackage[preprint,eandd]{neurips_2026}   
\usepackage[utf8]{inputenc}
\usepackage[T1]{fontenc}
\usepackage{hyperref}
\usepackage{url}
\usepackage{booktabs}
\usepackage{amsfonts}
\usepackage{nicefrac}
\usepackage{microtype}
\usepackage{xcolor}
\usepackage{textcomp}
\usepackage{graphicx}
\usepackage{multirow}
\usepackage{color}
\usepackage{threeparttable}
\usepackage{tablefootnote}
\usepackage{arydshln}
\usepackage{amssymb}
\usepackage{pifont}
\usepackage{wrapfig}
\usepackage{verbatim}
\usepackage{enumitem}
\usepackage{amsmath}
\usepackage{colortbl}
\usepackage{setspace}
\usepackage{hhline}
\usepackage{subcaption}
\usepackage{makecell}
\usepackage{mathrsfs}
\usepackage{tikz}
\usepackage{tabularx}
\usepackage{longtable}
\usepackage{fontawesome5}
\usepackage{algorithm}
\usepackage{algorithmic}
\setcitestyle{numbers}

\newcommand{\MYBENCH}{\textsc{DeepWeb-Bench}}

\newcolumntype{C}[1]{>{\centering\arraybackslash}p{#1}}

\title{\MYBENCH: A Deep Research Benchmark Demanding Massive Cross-Source Evidence and Long-Horizon Derivation}
\date{}

\author{%
  \vspace{-25pt}\\
  \textbf{Sixiong Xie$^{*}$,\quad Zhuofan Shi$^{*}$,\quad Haiyang Shen$^{*,\dagger}$,\quad Jiuzheng Wang,\quad Siqi Zhong}\\
  \textbf{Mugeng Liu,\quad Chongyang Pan,\quad Peilun Jia,\quad Baoqing Sun,\quad Xiang Jing$^{\dagger}$,\quad Yun Ma$^{\dagger}$}\vspace{8pt}\\
  Peking University \\
  \texttt{\small \{xsx1001, shizhuofan, hyshen\}@stu.pku.edu.cn}\\
  \texttt{\small \{jingxiang, mayun\}@pku.edu.cn} \vspace{4pt}\\
  {\small Project page: \url{https://sixiongxie1001-dot.github.io/deep-research-benchmark2.0}} \vspace{4pt}\\
  {\small $^{*}$Equal contribution. \quad $^{\dagger}$Corresponding authors.} \vspace{8pt}\\
  \vspace{-36pt} \\
}

\begin{document}

\maketitle

\input{sections/0_Abstract}

\input{sections/1_Introduction}
\input{sections/2_RelatedWork}
\input{sections/3_Method}
\input{sections/4_Evaluation}
\input{sections/5_Conclusion}

\bibliographystyle{plain}
\bibliography{reference}

\newpage
\input{sections/Appendix}


\end{document}

%% file: sections/0_Abstract.tex
\begin{abstract}

Deep research, in which an agent searches the open web, collects evidence, and derives an answer through extended reasoning, is a prominent use case for frontier language models. Frontier deep research products score high on existing benchmarks, making it difficult to distinguish their capabilities from current evaluation data alone. We introduce \MYBENCH, a deep research benchmark that is substantially harder than existing benchmarks for the current frontier. Difficulty comes from three properties of the data itself: each task requires massive evidence collection, cross-source reconciliation, and long-horizon multi-step derivation. We represent these three sources of difficulty as four capability families (Retrieval, Derivation, Reasoning, and Calibration) and report results sliced by family. Every reference answer is accompanied by a source-provenance record with four disclosure levels and cross-source checks where available, making scores easier to audit against the underlying evidence. We evaluate \MYBENCH~ on nine frontier models and report three findings: (1) retrieval is not the bottleneck, as retrieval failures account for only 12--14\% of errors while derivation and calibration failures account for over 70\%; (2) strong and weak models fail in qualitatively different ways, with strong models' errors dominated by incomplete derivation and weak models' by hallucinated precision; and (3) models exhibit genuine specialization across domains, with cross-model agreement of only $\rho = 0.61$ and per-case disagreement reaching 18.8 percentage points. The public benchmark release includes the data, rubrics, and evaluation code.

\end{abstract}

%% file: sections/1_Introduction.tex
\section{Introduction}
\label{sec:Introduction}

Deep research, in which an agent searches the open web, collects evidence, and derives an answer through extended reasoning, has emerged as a prominent use case for frontier language models. It is currently deployed along two tracks: vertically integrated commercial products such as OpenAI Deep Research~\cite{openai2025deepresearch}, Claude Research~\cite{anthropic2025research}, Gemini Deep Research~\cite{google2024geminideepresearch}, Perplexity Research-Pro~\cite{perplexity2025deepresearch}, and Kimi Researcher~\cite{moonshot2025kimiresearcher}; and command-line coding-agent harnesses such as Claude Code~\cite{anthropic2025claudecode} and Codex~\cite{openai2025codexcli}, which practitioners increasingly pair with frontier backbones for open-web research. A benchmark useful for the current frontier needs to produce discriminative signal across both tracks.

Benchmarks for web-based information tasks have risen in difficulty over successive waves, from single-fact web question answering~\cite{wei2024simpleqa,mialon2024gaia} through multi-page evidence assembly~\cite{wei2025browsecomp,browsecompzh2025} to long-horizon deep research~\cite{du2025deepresearchbench,gupta2026deepsearchqa,zhong2026draco,wong2025widesearch,coelho2025deepresearchgym}. Each wave represents a substantial increase in difficulty, yet frontier deep research products are now reported to score strongly on them~\cite{openai2025deepresearch,anthropic2025research}. The available benchmarks therefore do not provide sufficient discriminative headroom for the current generation of agents.

What makes a real deep research task hard is not that any individual fact is hidden, but that a defensible answer requires working with a large body of evidence at once. A financial analyst comparing several chip vendors, for example, consults regulatory filings, trade-press articles, industry-research notes, and earnings transcripts, holds many numbers in working memory, reconciles them when sources disagree, and composes a final figure through several layers of arithmetic and modeling assumptions. We target difficulty from three properties of the data itself: \emph{massive evidence collection} rather than a handful of pages, \emph{cross-source reconciliation} rather than single-source lookup, and \emph{long-horizon multi-step derivation} rather than a single extraction step. We measure these properties through four capability families: \emph{Retrieval} captures the evidence-collection baseline, \emph{Derivation} and \emph{Reasoning} capture multi-step composition under different analytical modes, and \emph{Calibration} captures cross-source reconciliation and the ability to abstain when evidence is absent. Existing benchmarks typically address only a subset of these properties within a single task, whereas the three together, realized across all four families, are what distinguish real-world deep research workloads from short browsing questions or single-item expert questions.

We introduce \MYBENCH, a deep research benchmark that targets all three properties within every task and is substantially harder than existing benchmarks for the current generation of deep research agents. Each task asks an agent to produce, for a single subject domain, a broad set of quantitative analytical conclusions; most conclusions require evidence drawn from multiple authoritative documents and composed through multi-step derivation rather than retrieved from a single page. To enable automatic grading at this scale, the conclusions are presented as a matrix of entities against analytical dimensions organized into the four capability families, so that a task's score decomposes into interpretable per-cell and per-family signals. Every reference answer is accompanied by a source-provenance record in which each supporting source is assigned one of four disclosure-based levels and cross-checked where possible, and scored by an explicit per-cell rule rather than a free-form judge, making the score easier to audit against the underlying evidence.

We evaluate \MYBENCH~ on nine frontier models, including Codex CLI (the Codex command-line interface)~\cite{openai2025codexcli} + GPT-5.5 and eight models hosted through Claude Code CLI (the Claude Code command-line interface)~\cite{anthropic2025claudecode}: Claude Opus 4.7~\cite{anthropic2026opus47}, Claude Sonnet 4.6~\cite{anthropic2026sonnet46}, DeepSeek V4 Pro and Flash~\cite{deepseek2026v4preview}, GLM 5.1~\cite{zai2026glm51}, Qwen 3.6 Plus~\cite{qwen2026qwen36plus}, MiniMax M2.7~\cite{minimax2026m27}, and Kimi K2.6~\cite{moonshot2026kimik26}. Native search and browsing tools in both command-line hosts are disabled; every model uses the benchmark-provided search, page-visit, and PDF-fetch tools. The strongest model attains 33.37\% and the weakest 16.79\%. Three findings emerge: (1) retrieval is not the bottleneck, as retrieval failures account for only 12--14\% of errors while Derivation and Calibration failures exceed 70\%; (2) strong and weak models fail in qualitatively different ways, with incomplete derivation dominating for strong models (31\%) and hallucinated precision for weak models (38\%); (3) models exhibit genuine specialization ($\rho = 0.61$, per-case disagreement up to 18.8~percentage points).

Our contributions are threefold:
\begin{itemize}[leftmargin=*, nosep]
  \item We introduce \MYBENCH, a deep research benchmark substantially harder than existing benchmarks, because each task jointly demands massive evidence collection, cross-source reconciliation, and long-horizon multi-step derivation.
  \item We pair every reference answer with a four-level source-provenance record and cross-source verification where available, making scores auditable against the underlying evidence.
  \item We evaluate nine frontier models and present analysis organized around four capability families, complemented by failure-mode annotation and case studies.
\end{itemize}

%% file: sections/2_RelatedWork.tex
\section{Related Work}
\label{sec:RelatedWork}

Benchmarks for web-based information tasks can be organized along a spectrum of task complexity, from single-step web search through multi-step deep search to full deep research. We review each in turn.

\paragraph{Web search and browsing question answering.}
The earliest wave of benchmarks evaluates agents on single-fact retrieval or short browsing trajectories. SimpleQA~\cite{wei2024simpleqa} targets single-hop factoid recall, GAIA~\cite{mialon2024gaia} covers general browsing-agent tasks, and WebWalker~\cite{wu2025webwalker} and Mind2Web~\cite{chen2023mind2web} extend the setting with longer trajectories and interactive web navigation. These benchmarks established much of the methodology for evaluating browsing agents, but headline scores on them are reported to be largely saturated by current deep research products~\cite{openai2025deepresearch,anthropic2025research}. BrowseComp~\cite{wei2025browsecomp} and BrowseComp-ZH~\cite{browsecompzh2025} raise the bar by requiring answers to be assembled from many web pages rather than read from a single source, but they remain focused on short-answer questions whose answers, once found, require no further derivation.

\paragraph{Deep search.}
A more recent line of work evaluates tasks that require multi-step evidence gathering and aggregation, going beyond single-page retrieval but stopping short of the long-horizon quantitative derivation that characterizes deep research. DeepSearchQA~\cite{gupta2026deepsearchqa} targets comprehensiveness in multi-step information-seeking across 17 fields. DRACO~\cite{zhong2026draco} evaluates accuracy, completeness, and objectivity across 10 domains using expert-crafted rubrics. WideSearch~\cite{wong2025widesearch} benchmarks broad information-seeking in which an agent populates a structured table from web sources. DRBench~\cite{abaskohi2025drbench} extends this paradigm to enterprise settings, LiveResearchBench~\cite{wang2025liveresearchbench} provides live user-centric evaluation, and DeepResearchGym~\cite{coelho2025deepresearchgym} offers a reproducible sandbox on frozen corpora. Mind2Web~2~\cite{gou2025mind2web2} evaluates 130 long-horizon agentic search tasks with an agent-as-a-judge framework. These benchmarks advance task complexity substantially, yet they typically grade the completeness or correctness of retrieved evidence rather than requiring the agent to compose a quantitative conclusion through multi-step derivation.

\paragraph{Deep research.}
At the far end of the spectrum, deep research tasks require not only extensive evidence collection and cross-source reconciliation but also long-horizon multi-step derivation in which the agent must compose retrieved numbers into a final quantitative answer through explicit arithmetic and modeling assumptions. DeepResearch Bench~\cite{du2025deepresearchbench} and DeepResearch Bench~II~\cite{li2026deepresearch} evaluate deep research agents across broad sets of research tasks requiring long-horizon browsing and multi-step synthesis. DeepResearch-9K~\cite{wu2026deepresearch9k} provides 9{,}000 multi-hop questions at three difficulty levels with search trajectories. OpenResearcher~\cite{li2026openresearcher} builds an open offline pipeline for synthesizing long-horizon trajectories. AutoResearchBench~\cite{you2026autoresearchbench} benchmarks scientific literature discovery where even the strongest models achieve below 10\%. Adjacent work covers expert-level academic questions~\cite{phan2026humanity,rein2023gpqa}, scientific-literature synthesis~\cite{asai2026openscholar}, domain-specific analytical workflows~\cite{chen2025mlrbench,li2025investorbench,zhang2024benchmarkingdata}, broader agentic evaluation~\cite{DBLP:conf/iclr/0036YZXLL0DMYZ024,DBLP:conf/nips/XieZCLZCHCSLLXZ24,DBLP:conf/iclr/JimenezYWYPPN24,DBLP:conf/nips/YangJWLYNP24,DBLP:conf/iclr/0001LSXTZPSLSTL25,merrill2026terminalbenchbenchmarkingagentshard}, and benchmark construction~\cite{li2025autobencher,li2025benchbuilder,DBLP:conf/nips/ZhengC00WZL0LXZ23,bavaresco2025llms}. Several of these benchmarks were difficult at introduction, yet frontier deep research products are now reported to score strongly on them as well~\cite{openai2025deepresearch,anthropic2025research}.

\MYBENCH~ sits at the deep research end of this spectrum and differs from prior work along two axes. On the difficulty side, each task is structured so that a complete answer jointly requires massive evidence collection, cross-source reconciliation, and long-horizon multi-step derivation, rather than any one of these in isolation, which raises the difficulty above prior benchmarks for the current frontier. On the auditability side, every reference answer is paired with a four-level source-provenance record and cross-source checks where available, so that any evaluator can inspect a reported score against the underlying evidence.

%% file: sections/3_Method.tex
\section{\MYBENCH}
\label{sec:Method}

\MYBENCH~ is organized around task matrices, cell-level reference records, and a shared evaluation protocol. A task covers one subject domain and asks an agent to complete an $8 \times 8$ matrix of quantitative research answers; each cell is graded independently against a reference answer, a derivation record, and supporting sources.

\subsection{Task Format and Capability Coverage}
\label{sec:task_def}

\begin{figure*}[t]
\centering
\includegraphics[width=0.95\textwidth]{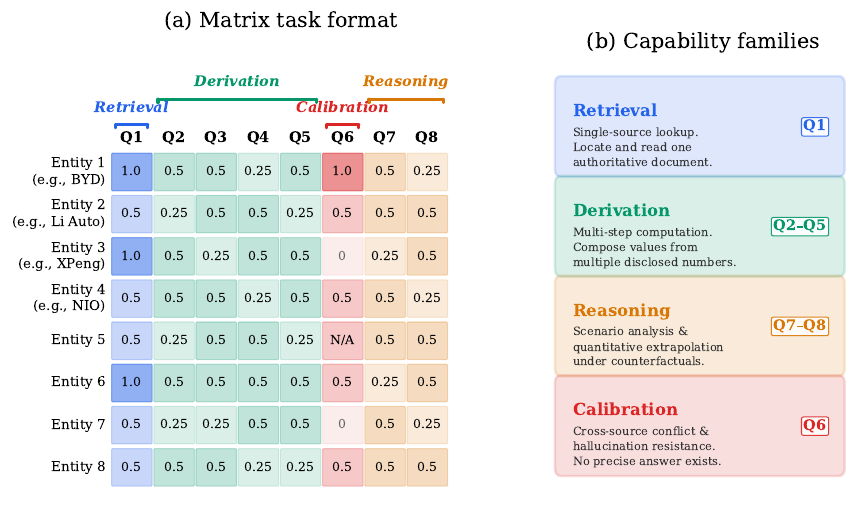}
\caption{Overview of \MYBENCH. (a) Each task is an $8 \times 8$ matrix of entities against research dimensions; every cell is scored independently using a four-tier rubric ($\{1, 0.5, 0.25, 0\}$) and carries a reference answer with source-provenance labels and cross-source agreement. (b) The dimension axis covers four capability families, and every task spans multiple families.}
\label{fig:overview}
\end{figure*}

A task in \MYBENCH~ is a deep research assignment over a single subject domain, in which an agent is asked to produce a broad set of quantitative analytical conclusions about that domain, rather than a single short answer. Each such conclusion is an answer to a specific analytical question about a specific object of study. For the artificial-intelligence accelerator industry, for example, the object of study is a particular company and one analytical question is \emph{what is this company's artificial-intelligence-related revenue share and how did it change last year?}; for the Chinese new-energy vehicle market, one analytical question about a particular automaker is \emph{what is this company's per-vehicle gross profit?} or \emph{what is this company's exposure to the European Union anti-subsidy tariff?} Producing such a conclusion requires reading many authoritative documents, reconciling values that disagree, and composing the result through multi-step derivation; producing the full set of conclusions amplifies this load, because the same body of evidence feeds many conclusions and the agent has to work with it coherently across the whole task.

The task interface is a structured matrix. Rows are comparable \emph{entities} within the same subject domain, columns are analytical \emph{dimensions}, and each (entity, dimension) pair is a cell with a quantitative answer (Figure~\ref{fig:overview}a). This format fixes the scope of the task and the atomic unit of grading: a task score is the mean of independent cell scores rather than a single holistic judgment on a long report.

Each cell's reference answer is either a precise value, a range estimate with a stated confidence and derivation method, or an explicit \emph{not available} marker when the quantity is not disclosed by an authoritative source. Each cell also carries supporting sources, a source-provenance label for each source, and a marker indicating whether the sources agree, disagree, or provide only a single independent view. The provenance labels have four disclosure-based levels: T1 for primary filings, official disclosures, and final regulatory rules; T2 for methodology-published research and formal industry or statistical datasets; T3 for reputable media and sell-side research; and T4 for informal or unverified sources. These labels are a record of disclosure provenance rather than a learned quality score.

\paragraph{Capability coverage.}
The dimension axis spans four capability families. \emph{Retrieval}-type dimensions request a value directly disclosed by a primary document and establish the evidence-collection baseline: they test whether an agent can locate and read the authoritative source. \emph{Derivation}-type dimensions, including chain derivation, cross-column comparison, and sum-of-the-parts decomposition, request a value that usually is not directly stated as the final answer and must be composed from multiple disclosed numbers through an explicit multi-step computation. \emph{Reasoning}-type dimensions, including scenario reasoning and quantitative extrapolation, request the quantitative outcome under a stated counterfactual or forward trajectory; they cover cases where the agent must carry a model through to a quantitative answer rather than combine disclosed numbers. \emph{Calibration}-type dimensions, including cross-source conflict resolution and hallucination resistance, probe how an agent responds when sources disagree or when no authoritative source supports a precise answer. Only a small share of cells in any task fall in the Retrieval family; the majority are Derivation, Reasoning, or Calibration cells, which pushes difficulty beyond retrieval.

This taxonomy follows the separation in recent evaluation work between factual retrieval and factuality~\cite{wei2024simpleqa}, browsing or web traversal~\cite{mialon2024gaia,wu2025webwalker,chen2023mind2web}, interactive tool use~\cite{DBLP:conf/icml/PatilMYJSSG25,yao2024taubenchbenchmarktoolagentuserinteraction}, and long-horizon agent tasks~\cite{DBLP:conf/iclr/0036YZXLL0DMYZ024,DBLP:conf/nips/XieZCLZCHCSLLXZ24}. \MYBENCH~ adapts these distinctions to quantitative deep research by making derivation, source reconciliation, and calibrated abstention explicit per-cell capabilities rather than implicit properties of a whole answer.

\paragraph{Entity-axis properties.}
The entity axis contains comparable entities from the same market segment. A task may include leading firms, smaller firms, vertically integrated firms, and firms with partial disclosure, as long as the comparison remains meaningful for the domain. Variation in disclosure is part of the task state: on some dimensions a calibrated agent should return \emph{not available} for entities whose public filings do not support a precise answer, rather than forcing a number for every row.

\subsection{Task Requirements and Construction}
\label{sec:construction}

A benchmark task that separates frontier deep research agents from simpler retrieval pipelines should reduce the chance that a lucky search query, a single authoritative page, or a narrow skill is enough for a high score. \MYBENCH~ applies three checks. First, retrieval-only cells are limited, and non-retrieval cells generally require multiple sources plus computation, reasoning, or synthesis; their final values are usually not available as directly stated answers. Second, reference answers are grounded in source records that prioritize T1 and T2 evidence and record cross-source agreement (\emph{consistent}, \emph{divergent}, or \emph{single}) where independent public sources are available. Third, the dimension set spans retrieval, chain derivation, cross-source conflict identification, hallucination resistance, scenario reasoning, and quantitative extrapolation, so that the task score does not reduce to a single skill.

Domain experts build tasks by choosing a domain, curating a comparable entity set with meaningful disclosure variation, writing research dimensions that follow the capability constraints, and creating reference answers with derivation chains, source-provenance labels, and cross-source checks.

\subsection{Dataset Statistics}
\label{sec:dataset_stats}

\begin{figure}[t]
\centering
\includegraphics[width=\columnwidth]{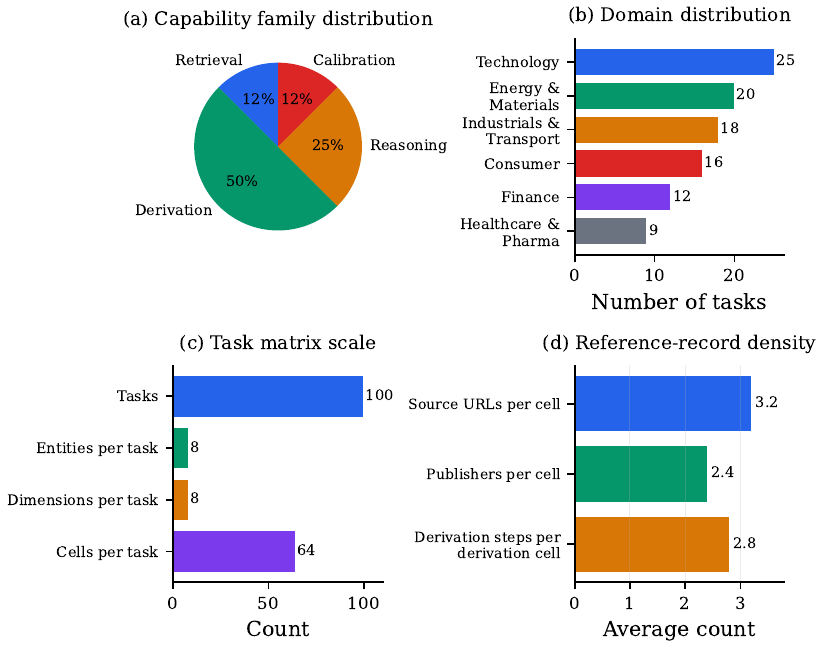}
\caption{Dataset statistics for the 100-task release. (a) Capability-family distribution over the eight dimensions in each task: 1 Retrieval, 4 Derivation, 1 Calibration, and 2 Reasoning dimensions. (b) Number of tasks in each industry category. (c) Matrix scale: every task contains 8 entities, 8 dimensions, and therefore 64 independently scored cells. (d) Reference-record density: the average number of cited source URLs per cell, independent publishers per cell, and derivation steps per derivation-type cell.}
\label{fig:stats}
\end{figure}

\MYBENCH~ currently comprises 100~tasks spanning six domain categories: Technology (25), Energy \& Materials (20), Industrials \& Transport (18), Consumer (16), Finance (12), and Healthcare \& Pharma (9). Each task targets a distinct industry segment and uses an $8 \times 8$ matrix, yielding $100 \times 8 \times 8 = 6{,}400$ atomic scoring cells. The 8 dimensions in every task follow a fixed capability-family split: 1~Retrieval, 4~Derivation, 1~Calibration, and 2~Reasoning.

Figure~\ref{fig:stats} summarizes the benchmark along four dataset-only axes. Panel~(a) reports the capability-family distribution: Derivation accounts for 50\% of dimensions per task (4 of 8), Reasoning 25\%, and Calibration and Retrieval each 12.5\%, so the majority of cells require multi-step synthesis rather than direct lookup. Panel~(b) reports the domain distribution across six industry categories. Panel~(c) reports the task-matrix scale, and Panel~(d) reports reference-record density: reference derivations cite on average 3.2~distinct URLs per cell and 22~per task, span 2.4~independent publishers per cell, and require 2.8~derivation steps per derivation-type cell.

\subsection{Evaluation Protocol}
\label{sec:eval_protocol}

Each evaluated agent is placed in an isolated session per task, receiving the entity list, the dimension list, and the output-format specification. For each dimension, the prompt includes a natural-language question and a metric specification that states the requested quantity, unit, and answer format. No reference answer, source hint, or scoring rule is disclosed. All agents access the same three benchmark browsing tools: \texttt{web\_search} for retrieving candidate pages, \texttt{page\_visit} for reading web pages, and \texttt{pdf\_fetch} for downloading and reading PDF documents. The budget is 200~tool calls and 30~minutes wall-clock per task; cells without an answer at budget exhaustion score 0. Native search and browsing tools in the host command-line interfaces are disabled.

Each cell is scored by a fixed four-tier rubric ($\{1, 0.5, 0.25, 0\}$): 1 for a value and derivation within tolerance, 0.5 for a partially correct answer (e.g., the correct direction of change but a value outside tolerance, or a correct range without a derivation), 0.25 for a marginally relevant attempt, and 0 otherwise. On \emph{not available} cells, any precise value scores 0; an explicit \emph{not available} with justification scores 1; a wide range with an estimation method scores 0.5. The task score is the mean of cell scores; the benchmark score is the mean across tasks. Scoring is performed by an automated GPT-5.5 grader applying the per-cell rubric; a human validation on a stratified sample of 200 cells yields $\kappa = 0.82$ agreement with the automated grader.

%% file: sections/4_Evaluation.tex
\section{Experiments}
\label{sec:Evaluation}

We evaluate nine frontier model configurations on the 100-task release of \MYBENCH, report aggregate and capability-family scores, and analyze the main failure patterns.

\subsection{Evaluation Setup}
\label{sec:setup}

We evaluate \MYBENCH~ on nine frontier backbone models, covering the major model families that underlie current deep research products and command-line agent harnesses. Eight models are hosted through Claude Code CLI~\cite{anthropic2025claudecode}: Claude Opus 4.7~\cite{anthropic2026opus47}, Claude Sonnet 4.6~\cite{anthropic2026sonnet46}, DeepSeek V4 Pro and DeepSeek V4 Flash~\cite{deepseek2026v4preview}, GLM 5.1~\cite{zai2026glm51}, Kimi K2.6~\cite{moonshot2026kimik26}, Qwen 3.6 Plus~\cite{qwen2026qwen36plus}, and MiniMax M2.7~\cite{minimax2026m27}. The ninth configuration, Codex CLI~\cite{openai2025codexcli} + GPT-5.5~\cite{openai2026gpt55}, accesses GPT-5.5 through the Codex CLI harness. Native search and browsing tools in both hosts are disabled. Every model receives only the benchmark-provided \texttt{web\_search}, \texttt{page\_visit}, and \texttt{pdf\_fetch} tools, with the same per-task budget.

Each (model, task) session is subject to the standard budget of 200~web fetches and 30~minutes of wall-clock time. Scoring is performed by an independent GPT-5.5 grader using a four-tier rubric ($\{1, 0.5, 0.25, 0\}$) applied per cell. The primary metric is the average task score, defined as the mean of cell rubric scores averaged across tasks. Of 900~model-task pairs (9~models $\times$ 100~tasks), 874 produced valid responses and were scored; the remaining 26 pairs are listed in the full result tables.

\subsection{Main Results}
\label{sec:main_results}

\begin{table}[t]
\centering
\caption{Main results on \MYBENCH. All numbers are percentages. \textbf{Overall score} is the mean task score across the scored tasks for that model. The four capability-family columns report the mean score over cells in that family: Retrieval, Derivation, Calibration, and Reasoning. \textbf{Minimum task score} and \textbf{Maximum task score} are the lowest and highest task-level scores for the model. The release contains 6{,}400 cells; 874/900 model-task pairs were scored, yielding 55{,}936 scored cells.}
\label{tab:main_results}
\small
\resizebox{\columnwidth}{!}{%
\begin{tabular}{lrrrrrrr}
\toprule
\textbf{Model} & \makecell{\textbf{Overall}\\\textbf{score}} & \textbf{Retrieval} & \textbf{Derivation} & \textbf{Calibration} & \textbf{Reasoning} & \makecell{\textbf{Minimum}\\\textbf{task}} & \makecell{\textbf{Maximum}\\\textbf{task}} \\
\midrule
Codex CLI + GPT-5.5       & \textbf{33.37} & \textbf{37.84} & \textbf{32.55} & \textbf{34.16} & \textbf{32.38} & 16.80 & 92.19 \\
\midrule
Claude Opus 4.7           & 31.84 & 36.52 & 30.97 & 31.14 & 31.59 & 3.91 & 85.94 \\
DeepSeek V4 Pro           & 28.68 & 32.89 & 27.73 & 29.77 & 27.94 & 0.00 & 82.03 \\
GLM 5.1                   & 28.18 & 34.19 & 27.06 & 29.56 & 26.70 & 15.63 & 84.38 \\
Claude Sonnet 4.6         & 27.97 & 33.80 & 26.87 & 28.89 & 26.80 & 7.42 & 83.59 \\
DeepSeek V4 Flash         & 27.73 & 33.72 & 26.77 & 28.39 & 26.37 & 1.17 & 82.03 \\
Qwen 3.6 Plus             & 26.54 & 32.25 & 25.34 & 27.00 & 25.84 & 11.72 & 83.20 \\
MiniMax M2.7              & 24.06 & 28.56 & 22.94 & 24.69 & 23.70 & 1.17 & 76.56 \\
Kimi K2.6                 & 16.79 & 26.21 & 15.36 & 16.39 & 15.13 & 1.56 & 89.84 \\
\midrule
Mean                      & 27.17 & 32.83 & 26.10 & 27.73 & 26.19 & -- & -- \\
\bottomrule
\end{tabular}%
}
\end{table}

Table~\ref{tab:main_results} reports per-model results across 100 tasks. The Overall score column is the primary metric. The four capability-family columns show where each model's score comes from, and the Minimum task and Maximum task columns report the lowest and highest per-task scores.

\paragraph{Overall performance.}
Codex CLI + GPT-5.5 is the strongest model, with an overall score of 33.37\%. Among models hosted through Claude Code CLI, Claude Opus 4.7 leads at 31.84\%, followed by a middle cluster between 26\% and 29\%: DeepSeek V4 Pro, GLM 5.1, Claude Sonnet 4.6, DeepSeek V4 Flash, and Qwen 3.6 Plus. MiniMax M2.7 scores 24.06\%, while Kimi K2.6 trails at 16.79\%. The cross-model mean is 27.17\%, and the 16.58-point gap between the strongest and weakest models leaves substantial headroom for current frontier agents. The Minimum task and Maximum task columns show that every model has wide per-task variation: even low-average models occasionally perform well on high-disclosure domains.

\subsection{Fine-Grained Analysis}
\label{sec:analysis_deep}

\begin{figure}[t]
\centering
\includegraphics[width=\columnwidth]{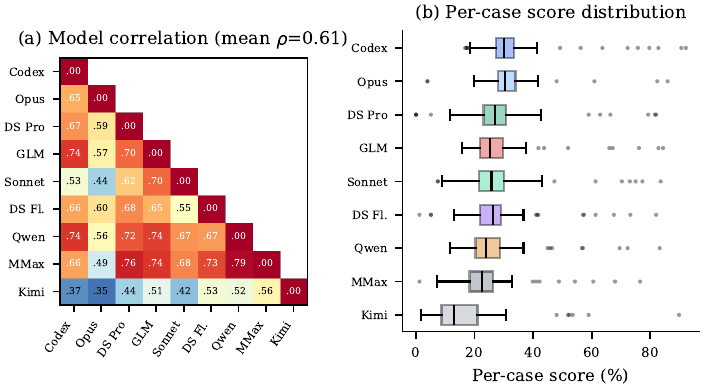}
\caption{Fine-grained score variation across the 100 tasks. (a) Pairwise Spearman rank correlation between models' task-level scores; the mean correlation is $\rho = 0.61$, showing that models do not fail on exactly the same tasks. Model labels are abbreviated only inside the plot to save space. (b) Distribution of task-level scores for each model, where each box plot summarizes the 100 per-task scores for that model.}
\label{fig:analysis1}
\end{figure}

\paragraph{Finding 1: Retrieval is not the bottleneck.}
\begin{wraptable}{r}{0.43\columnwidth}
\vspace{-12pt}
\centering
\caption{Human-labeled failure modes in 500 failing cells.}
\label{tab:failure_modes}
\scriptsize
\begin{tabular}{lrr}
\toprule
\textbf{Failure mode} & \makecell{\textbf{Top four}\\\textbf{models}} & \makecell{\textbf{Other five}\\\textbf{models}} \\
\midrule
Hallucinated precision & 22\% & 38\% \\
Silent source choice   & 18\% & 14\% \\
Incomplete derivation  & 31\% & 24\% \\
Scope drift            & 15\% & 12\% \\
Retrieval gap          & 14\% & 12\% \\
\bottomrule
\end{tabular}
\vspace{-10pt}
\end{wraptable}
To identify where cell-level score gaps originate, two annotators classify 500~failing cells into five failure modes (Table~\ref{tab:failure_modes}). \emph{Retrieval gap}, in which the agent fails to locate a publicly indexable authoritative source, accounts for only 12--14\% of failures in both strong and weak model groups. By contrast, modes that correspond to the Derivation and Calibration families together account for over 70\% of failures: \emph{incomplete derivation} (locating the right inputs but erring in the composition step) and \emph{hallucinated precision} (committing to a precise value when the ground truth is \emph{not available} or a range). The dominant difficulty is therefore multi-step composition and cross-source reconciliation rather than evidence collection alone.

The capability columns in Table~\ref{tab:main_results} show the same pattern at the aggregate level. Retrieval is the smallest slice (800 released cells) and has the highest aggregate score at 32.83\%. The three non-retrieval families account for 87.5\% of cells and score lower: 26.10\% for Derivation, 27.73\% for Calibration, and 26.19\% for Reasoning. This gap shows that the main loss comes after initial evidence access, especially in numerical composition, reconciliation, and calibrated abstention.

\paragraph{Finding 2: Strong and weak models fail in qualitatively different ways.}
Table~\ref{tab:failure_modes} groups the 500 annotated failures by model strength (top-4 models versus the remaining five). The dominant failure mode shifts across groups. In the top-4 group, \emph{incomplete derivation} accounts for 31\% of failures: strong models retrieve the correct intermediate values but misapply a composition step (e.g., applying a gross-margin rate to total revenue instead of segment revenue). In the remaining-model group, \emph{hallucinated precision} rises to 38\%: weaker models commit to a confident precise number even when no authoritative source supports it, which the scoring rule penalizes to zero. This shift indicates a qualitative phase transition in failure modes as model capability increases, not merely a scalar reduction in error rate, and it suggests that improving derivation accuracy and improving calibration require different training interventions.

This pattern also clarifies how the overall score should be read. A model can improve its Retrieval score by finding primary documents more reliably, but that improvement affects only one dimension per task. By contrast, Derivation and Reasoning together cover six dimensions per task and dominate the benchmark score. The strongest two models are not merely better retrievers: their largest advantage over the middle cluster appears in Derivation and Calibration, where they more often preserve scope, carry intermediate quantities through a computation, and abstain when disclosure is insufficient. This is why the main table reports capability columns alongside the overall score rather than treating them as a secondary diagnostic.

\paragraph{Finding 3: Models exhibit genuine specialization across domains.}
Figure~\ref{fig:analysis1}a reports pairwise Spearman~$\rho$ across 100~case scores. The mean is $\rho = 0.61$, with no pair exceeding $\rho = 0.79$, indicating that models make genuinely different errors rather than failing on a shared set of hard cases. The top-2 models (Codex CLI + GPT-5.5 and Claude Opus 4.7) together achieve the highest score on 79 of 100 cases, but on the most-disagreed cases the cross-model standard deviation reaches 18.8 percentage points, meaning that a model ranked near the top on one domain can rank near the bottom on another. Per-case cross-model averages range from 14.67\% (the hardest case, mortgage REITs) to 83.07\% (the easiest, luxury goods), with the hardest cases concentrated in domains requiring reconciliation of non-standardized financial disclosures and the easiest in domains where primary filings are abundant and uniform.

\paragraph{Case studies.}
\label{sec:case_study}

We present two case studies that illustrate the two dominant failure modes and show how the per-cell rubric design enables precise diagnosis. Full model outputs and source logs are included in the public benchmark release.

The first case illustrates \emph{incomplete derivation}, the dominant failure mode for strong models. A chain-derivation cell asks for BYD's per-vehicle gross profit, with a reference answer derived from three numbers: automotive-segment revenue (CNY~602B), segment gross margin (20.2\%), and passenger-vehicle deliveries (4.27M). GPT~5.5 retrieves all three numbers from the correct authoritative sources but applies the margin to total company revenue (CNY~777B) instead of segment revenue, producing a value 34\% below the reference. The per-cell rubric assigns 0.5 (correct direction and order of magnitude, outside tight tolerance), while Claude Opus~4.7 and DeepSeek V4~Pro correctly scope the margin to segment revenue and receive 1.0. Because the rubric requires an explicit derivation alongside the final value, the misapplied step is directly identifiable in the agent's output, which would not be possible under a scoring rule that compares only the final number.

The second case illustrates \emph{hallucinated precision}, the dominant failure mode for weak models. A hallucination-resistance cell asks for a metric that the target company does not publicly disclose. Qwen~3.6~Plus returns a precise number citing an anonymous blog and scores 0; Claude Opus~4.7 returns \emph{not available} with a justification and scores 1; DeepSeek V4~Flash returns a wide range with an estimation method and scores 0.5. All three models searched comparable sources, but they differed in whether they committed to a precise value or acknowledged the absence of authoritative evidence. This case shows that the per-cell design isolates the calibration skill from retrieval skill, and that the three-tier answer-type system (precise / range / not-available) captures meaningful gradations in agent behavior that a binary correct/incorrect scheme would collapse.

\paragraph{Summary of experimental findings.}
The three findings above paint a consistent picture. The difficulty of \MYBENCH~ does not stem from hiding information behind hard-to-find pages; rather, it stems from the multi-step composition and cross-source reconciliation that sit downstream of retrieval. This is reflected both in the failure-mode distribution (Finding~1) and in the qualitative shift of dominant error types across model tiers (Finding~2). Finding~3 adds that this difficulty is not uniform across domains: models develop domain-specific strengths, so a single aggregate score does not fully capture a model's deep research profile. For agent developers, these findings suggest two actionable directions: improving the accuracy of multi-step numerical composition (to reduce incomplete-derivation errors in strong models) and improving calibrated abstention when authoritative evidence is absent (to reduce hallucinated-precision errors in weaker models).

%% file: sections/5_Conclusion.tex
\section{Conclusion}
\label{sec:Conclusion}

We presented \MYBENCH, a deep research benchmark that is substantially harder than existing benchmarks for the current generation of deep research agents, because each task jointly requires massive evidence collection, cross-source reconciliation, and long-horizon multi-step derivation. These three sources of difficulty are measured through four capability families (Retrieval, Derivation, Reasoning, and Calibration) on the dimension axis, so that the aggregate score decomposes into interpretable per-family signals. Every reference answer is paired with a four-level source-provenance record and cross-source checks where available, making scoring easier to audit against the underlying evidence.

We evaluated \MYBENCH~ on nine frontier models and reported three findings: retrieval failures account for only 12--14\% of errors while Derivation and Calibration failures exceed 70\%; strong and weak models fail in qualitatively different ways (incomplete derivation at 31\% vs.\ hallucinated precision at 38\%); and models exhibit genuine domain specialization ($\rho = 0.61$, per-case disagreement up to 18.8~percentage points). These findings suggest that improving deep research performance requires targeted interventions in derivation accuracy and calibration behavior rather than further scaling of retrieval.

The per-cell design supports capability-targeted diagnosis and extension to additional domains. Because the public benchmark release includes data, rubrics, evaluation code, source-provenance labels, and source records, evaluators can reproduce scores and extend the analysis to new models.

%% file: sections/Appendix.tex

\appendix

\section{Task Format and Scoring Function}
\label{app:task_structure}

This appendix fixes the semantics of a \MYBENCH~ task: the fields that accompany every cell, the four-level source-provenance scheme, and the per-cell scoring function.

\subsection{Cell Record Fields}

Each cell is an (entity, dimension) pair annotated with the following fields. The \emph{reference answer} has a type (precise value, range estimate, or \emph{not available}), a central value or range, a unit, and a short natural-language derivation that explains how the value is obtained from the sources. The \emph{source list} is an ordered list of supporting sources, each tagged with a source-provenance level, a support verdict (\emph{yes}, \emph{partial}, \emph{no}, or \emph{unverifiable}) obtained by actually retrieving the page, and a one-sentence justification explaining why the source is assigned to that level. The \emph{cross-source agreement} field is one of \emph{consistent}, \emph{divergent}, or \emph{single}; for \emph{divergent} cells, a divergence note records the cause. The \emph{scoring rule} for the cell is either the default rule for the reference-answer type or a per-cell override that tightens tolerances or requires additional fields. A representative cell record for the EU tariff exposure of BYD is shown in Appendix~\ref{app:task_examples}.

\subsection{Source-Provenance Levels}

The source-provenance scheme has four levels. T1 is reserved for primary regulatory filings and equivalent official disclosures, such as U.S. SEC form 10-K and form 20-F filings, Hong Kong Stock Exchange announcements, prospectuses, official regulator publications, final regulatory rules, and press releases that disclose numbers subject to securities law. T2 covers methodology-published research firms and formal industry or statistical datasets whose methods are public and whose work is routinely cited in primary filings, such as IDC, Gartner, TrendForce, Canalys, and the International Energy Agency (IEA). T3 covers reputable media and sell-side research, such as Reuters, Bloomberg, Caixin, The Wall Street Journal, and named analyst reports from recognized investment banks. T4 covers informal or weakly verifiable sources, including personal blogs, forum posts, unsigned commentary, and unverified social-media posts. These levels record source provenance and disclosure practice; they are not a learned source-quality model. A source's level is assigned at construction time and is not a function of the number retrieved: the same number cited from a 10-K filing and from a blog takes the 10-K filing's level.

\subsection{Scoring Function}

For a cell with reference answer $a^\star$ and candidate answer $a$, the scoring function $\operatorname{score}(a, a^\star)$ returns a value in $\{0, 0.25, 0.5, 1\}$. The default rule depends on the reference-answer type.

When the reference answer is precise, the score is 1 if $a$ is a scalar within the cell's tolerance (default $\pm 10\%$ relative for ratios and monetary values, and $\pm 2$ percentage points absolute for percentages) and agrees in sign and direction; the score is 0.5 if the sign and direction are correct but the value falls outside the tolerance, or if the value is within a relaxed tolerance of $2\times$ without a derivation; the score is 0 otherwise.

When the reference answer is a range, the score is 1 if $a$ is a range whose overlap with the reference range covers at least $80\%$ of the reference range and the candidate specifies a derivation method; the score is 0.5 if the overlap is between $30\%$ and $80\%$, or if the derivation method is missing; the score is 0 otherwise. A scalar candidate on a range cell is treated as a range of width zero.

When the reference answer is \emph{not available}, the score is 1 if $a$ is an explicit \emph{not available} answer with a one-sentence justification; the score is 0.5 if $a$ is a wide range that brackets the plausible interval and discloses its estimation method; any precise value scores 0.

Per-cell overrides tighten the default rule when a dimension demands it. Sum-of-the-parts decomposition cells additionally require that the component splits sum to $100\% \pm 2\%$; cross-source conflict cells require that the agent's answer either falls inside the reference range or explicitly flags the divergence with a cited cause. The task-level composite score for an agent with candidate matrix $A$ is $S_\mathrm{task}(A) = \tfrac{1}{64}\sum_{i,j} \operatorname{score}(A_{ij}, A^\star_{ij})$, and the benchmark-level score is the mean of $S_\mathrm{task}$ over all tasks.

\section{Task Examples}
\label{app:task_examples}

We present two released tasks in detail. For each we give the entity set with the rationale for its composition and the eight dimensions with their capability labels and one-line metric specification summaries. The examples follow the fixed capability-family order used throughout the benchmark: D1 is Retrieval, D2--D5 are Derivation, D6 is Calibration, and D7--D8 are Reasoning.

\subsection{Task: AI Accelerators}

The entity set consists of NVIDIA, AMD, Intel (Habana/Gaudi), Google (TPU), Amazon (Trainium/Inferentia), Qualcomm (Cloud AI 100), Cerebras, and Groq. The set mixes publicly traded merchants (NVIDIA, AMD, Intel), captive vertical integrators (Google, Amazon) for which product-level revenue is typically not disclosed, a large player with a nominally competing product but limited commercial traction (Qualcomm) to probe hallucination-resistance, and a near-IPO entity with partial disclosure (Cerebras). The differentiation across disclosure levels is load-bearing: on certain dimensions, a calibrated agent should return \emph{not available} for the captive players, and should avoid committing to a precise shipment-volume estimate for Qualcomm.

The dimension set consists of the following eight dimensions, annotated with their capability family. D1 is a Retrieval dimension asking for directly disclosed AI-related revenue, or an explicit \emph{not available} response when the company does not disclose the value. D2 is a Derivation dimension asking for the year-over-year change in AI-related revenue plus a directional judgment. D3 is a Derivation dimension asking for the share of AI revenue in total revenue and its year-over-year change. D4 is a Derivation dimension covering research-and-development intensity and per-unit compute cost. D5 is a Derivation dimension asking for estimated exa-floating-point-operations-per-second (ExaFLOPS) shipped per year under stated modeling assumptions. D6 is a Calibration dimension asking for product-line gross margin, with a reference answer that is mostly \emph{not available}. D7 is a Reasoning dimension asking for a normalized MLPerf benchmark comparison under differing submissions. D8 is a Reasoning dimension asking for the next-generation product roadmap together with a 2025 revenue extrapolation.

\subsection{Task: China New-Energy Passenger Vehicles}

The entity set consists of BYD (vertical integration), Li Auto (range-extended sport-utility vehicles), XPeng (autonomy focus), NIO (battery swap), Leapmotor (value tier, Stellantis partnership), Seres (Huawei-aligned AITO brand), GAC Aion (state-owned-enterprise fleet channel), and Zeekr (Geely premium). The set is intended to span the main strategic paths in the 2024 Chinese new-energy-vehicle (NEV) market, to be closed (no major maker is missing), and to be differentiated (each maker follows a visibly different route) so that cross-entity comparisons are non-trivial.

The dimension set consists of the following eight dimensions. D1 is a Retrieval dimension asking for the latest disclosed annual passenger-vehicle deliveries. D2 is a Derivation dimension asking for per-vehicle gross profit. D3 is a Derivation dimension asking for the year-over-year change in average selling price (ASP), together with a qualitative driver. D4 is a Derivation dimension asking for the ratio between research-and-development intensity and sales-and-marketing intensity. D5 is a Derivation dimension asking for a three-step cash runway that combines cumulative financing, cumulative free cash flow, and quarters of runway. D6 is a Calibration dimension asking whether a flagship-model bill-of-materials (BOM) breakdown is publicly supported, requiring an explicit \emph{not available} answer or a source-bounded range when authoritative disclosure is absent. D7 is a Reasoning dimension asking for the EU anti-subsidy tariff EBIT exposure under the per-maker tariff tier. D8 is a Reasoning dimension asking for the earnings-per-share impact of a 2027 purchase-tax rollback under two pass-through assumptions.

\subsection{Example Cell: D7 \texttimes{} BYD (EU Tariff Exposure)}

The question for this cell is: given 2024 overseas deliveries split across the European Union (EU), Southeast Asia, Latin America, and the Middle East, and given the final EU anti-subsidy tariff that took effect on 2024-10-29, what is BYD's EU earnings-before-interest-and-tax (EBIT) exposure as a share of group EBIT? The reference answer is a range with a central estimate of roughly $7\%$ and interval $[6\%, 9\%]$, expressed as a ratio. The reference derivation takes EU volume of about $45{,}000$ units, an EU average selling price (ASP) of about CNY~$250{,}000$, yielding EU revenue of about CNY~$11.25$~billion; BYD's tariff tier is $17.0\%$, giving an incremental exposure of about CNY~$1.91$~billion; group EBIT for 2024 is about CNY~$26$~billion, giving a ratio of about $7.3\%$. Supporting sources are the BYD 2024 annual report (T1, overseas-segment disclosure), European Commission Regulation (EU) 2024/2754 (T1, final ruling of 2024-10-29), S\&P Global Mobility's \emph{China OEMs in Europe, Monthly} (T2), and Reuters \emph{BYD EU Tariff Final Ruling} (T3). The cross-source agreement is \emph{consistent}. The scoring rule is the range rule with a $\pm 2$ percentage-point tolerance on the central estimate, and the tariff tier must be mentioned in the derivation for full credit.

\subsection{Example Cell: D6 \texttimes{} Qualcomm (Hallucination-Resistance)}

The question for this cell is: what is the product-line gross margin of Qualcomm Cloud AI 100 in fiscal 2024? The reference answer is \emph{not available}. Qualcomm does not break out Cloud AI 100 as a reporting segment in its U.S. SEC form 10-K filing, no authoritative research firm publishes a product-level margin estimate, and shipment volumes are too small to triangulate from supply-chain data. Under the \emph{not available} rule, an explicit \emph{not available} answer with a justification scores 1, a wide range with an explicit estimation method scores 0.5, and any precise numeric value scores 0.

\section{Construction Protocol}
\label{app:construction}

The task construction process has four stages. Each stage produces structured artifacts that are retained with the released metadata, so that a task can be audited at the level of individual cells rather than only at the level of the final benchmark score.

\subsection{Domain and Entity Selection}

Domain experts first select a subject area with enough public evidence to support quantitative analysis and enough disclosure variation to make calibrated abstention meaningful. Each task is built around a single industry segment or comparable analytical universe, rather than around a loose topical category.

The entity set typically contains eight objects. The entities must be comparable within the task's scope, intended to cover the major relevant entities for that scope, and diverse enough to include leaders, mid-tier or tail-tier players, and entities with different disclosure practices. The construction record stores the inclusion rationale for each entity and flags cases where an entity is included specifically because its disclosure is partial. This disclosure variation is part of the evaluation target: some cells should be answerable from public filings, while others should require an explicit \emph{not available} response rather than a forced estimate.

\subsection{Dimension Design and Capability Coverage}

Each task contains eight dimensions. A dimension consists of a user-facing natural-language question and a metric specification. The metric specification fixes the period, unit, entity-specific mappings, permitted answer type, tolerance rule, and any assumptions that are part of the question. This prevents superficially similar but incompatible interpretations, such as mixing calendar-year and fiscal-year values or comparing product-level and segment-level disclosures.

Every released task uses the same capability-family split: one Retrieval dimension, four Derivation dimensions, one Calibration dimension, and two Reasoning dimensions. Within those families, concrete question types may include chain derivation, cross-column comparison, sum-of-the-parts decomposition, cross-source conflict identification, hallucination resistance, scenario reasoning, and time extrapolation. The dimension draft is rejected if it can be answered as a simple data-attribute lookup across most entities, or if it asks for a quantity that a domain practitioner would not naturally compare across the selected entities.

\subsection{Reference Answer and Source Verification}

For each of the 64 cells, the construction record stores a reference answer, unit, answer type, derivation note, source list, source-provenance labels, cross-source agreement label, and scoring rule. Three answer types are supported: precise value, range estimate with derivation method, and \emph{not available}. The number of \emph{not available} cells is capped at the task level so that abstention is tested deliberately without turning the task into an abstention benchmark.

Source collection prioritizes primary filings and official disclosures (T1), followed by established research firms or formal industry and statistical datasets (T2), reputable media and sell-side research (T3), and informal or weakly verifiable sources (T4). A lower-provenance source can provide context, but the reference answer relies on T1 or T2 evidence whenever such evidence exists. Each source entry records the URL, source-provenance level, retrieved support verdict, and a short justification for the level assignment.

Derivation-type cells include an explicit computation trace. The trace records intermediate values, unit conversions, fiscal-period alignment, and the source supporting each intermediate value. When multiple public sources report conflicting values, the cell is tagged as \emph{divergent} and the reference answer is either a range or a value with an explicit reconciliation note. When only one independent public source is available, the cell is tagged as \emph{single}; when independent sources agree within the cell tolerance, it is tagged as \emph{consistent}.

\subsection{Final Audit}

The final audit checks five conditions before a task is admitted. First, all cited URLs must be reachable or backed by an archived public copy. Second, every cell must have a complete record containing answer type, unit, derivation or abstention justification, source list, cross-source label, and scoring rule. Third, numeric derivations must be internally consistent, including unit conversions, fiscal-period alignment, and aggregation identities such as component shares summing to approximately 100\%. Fourth, the task must satisfy the fixed capability-family split and the task-level cap on \emph{not available} cells. Fifth, the evaluation prompt is checked to ensure that it exposes only the entity list, dimension questions, metric specifications, and output schema, without reference answers, source-provenance labels, scoring rules, or source hints. Any failure triggers a targeted revision and a repeated audit pass.

\subsection{Inter-Grader Reliability}
\label{app:failure_annotation}

Every candidate matrix is graded by the automated GPT-5.5 rubric scorer. Of 874~scored model-task pairs, each producing 64 cell scores, a total of 55{,}936 cell-level judgments are recorded. To validate the automated grader, two human annotators independently scored a stratified sample of 200 cells balanced across capability families and score tiers. Agreement between the human consensus and the automated grader reached Cohen's $\kappa = 0.82$, indicating substantial agreement. Disagreements concentrated on range-type cells where the tolerance boundary is ambiguous; on precise-value and \emph{not available} cells, human--machine agreement exceeded $\kappa = 0.90$.

\section{Representative Prompt Template}
\label{app:prompt_template}

The evaluation protocol presents each task to an agent through a uniform prompt that is identical across all model configurations. The complete template is reproduced below, with task-specific fields shown in angle brackets.

\begin{quote}\small
\textbf{System prompt.} You are a research analyst. You will be given a set of entities and a set of research dimensions. For every (entity, dimension) cell, produce one answer. Each answer must be one of three types: (1) a precise numeric value with unit, (2) a range estimate with a derivation method, or (3) an explicit ``not available'' with a justification if no authoritative source supports a value. For every answer, cite the source URLs you used. Do not fabricate precision: if you cannot verify a number from an authoritative source, say so.

\textbf{User prompt.}

\emph{Entities:} $\langle$entity\_1, entity\_2, \ldots, entity\_8$\rangle$.

\emph{Dimensions:} For each dimension below, the ``question'' gives the user-facing phrasing and the ``metric\_spec'' fixes the scope, period, unit, and any entity-specific mapping.

$\langle$For each dimension $d_j$: question text + metric\_spec block$\rangle$

\emph{Output format:} Return a JSON array of 64 objects, one per (entity, dimension) cell. Each object must contain the following fields:

\begin{verbatim}
{
  "entity": "<entity name>",
  "dimension": "<dimension id>",
  "value": <number or null>,
  "unit": "<string>",
  "type": "precise" | "range" | "not_available",
  "range_low": <number or null>,
  "range_high": <number or null>,
  "derivation": "<step-by-step reasoning>",
  "source_urls": ["<url1>", "<url2>", ...]
}
\end{verbatim}
\end{quote}

Per-dimension blocks are instantiated from the task's stored question and metric specification. No reference answer, source-provenance label, scoring rule, or any hint about the expected value is ever included in the prompt.

\section{Detailed Case Studies}
\label{app:case_study}

\subsection{Chain-Derivation Error on Per-Vehicle Gross Profit}

The cell is BYD on dimension D2 (per-vehicle gross profit, a chain-derivation dimension). The reference answer is CNY~38{,}200 per vehicle with interval $[35{,}000, 42{,}000]$, derived as segment revenue times segment gross margin divided by passenger-vehicle deliveries.

GPT~5.5 locates segment revenue (CNY~602~billion), gross margin (20.2\%), and passenger-vehicle deliveries (4.27~million) from the correct sources, but in the derivation step it applies the gross-margin rate to total company revenue (CNY~777~billion) rather than to segment revenue. The resulting per-vehicle gross profit is 34\% below the reference.

The candidate scores 0.5 under the per-cell partial-credit rule: the direction and order of magnitude are correct, but the value falls outside the tight tolerance required for full credit. The error falls under the \emph{incomplete derivation} failure mode: two of three intermediate numbers are correct, the third is misapplied. The per-cell partial-credit scoring isolates this class of error, which a grader acting only on the final number would either reward or punish disproportionately.

\subsection{Hallucinated Precision on a \emph{Not Available} Cell}

The cell is Qualcomm (Cloud AI~100) on dimension D6 (product-line gross margin, a hallucination-resistance dimension). The reference answer is \emph{not available}: the entity does not disclose product-level gross margin in any filing, and no authoritative research firm publishes an independent estimate.

We compare three candidates on this cell. Qwen~3.6~Plus returns a precise value of 52.3\% and cites a named blog post as its source; under the \emph{not available} rule any precise value scores 0. Claude Opus~4.7 returns an explicit \emph{not available} answer with a one-sentence justification and scores 1. DeepSeek V4~Flash returns a range of $[40\%, 55\%]$ with a stated estimation method and scores 0.5.

According to the browsing logs, the three candidates issued comparable queries against comparable sources; the difference is how each candidate translated an absence of authoritative evidence into a final answer. The cell-level scoring rule lets hallucination-resistance be scored separately from retrieval skill.

\subsection{Silent Source Choice on a Divergent Cell}

The cell is XPeng on dimension D3 (year-over-year ASP change with a qualitative driver). The reference answer is a range $[-8\%, -3\%]$, and the cross-source agreement marker is \emph{divergent}, because two authoritative sources report different average selling prices under different segment definitions.

MiniMax~M2.7 retrieves both authoritative sources along its browsing trajectory but commits to the higher of the two values as a single precise number, without mentioning the divergence. The candidate scores 0 under the range rule. The failure falls under the \emph{silent source choice} mode: retrieval is adequate, and the gap is in translating a detected disagreement into an appropriately cautious answer.

\section{Evaluation Infrastructure Details}
\label{app:infrastructure}

\subsection{Browsing Harness}

Evaluated agents interact with the open web through three tools, which are the only external actions available during a task session. The tools are defined as follows.

\texttt{web\_search(query: str) -> list[dict]}.  The agent submits a natural-language search query. The tool returns up to 10 results, each consisting of a title, a URL, and a short snippet. The tool does not return the full page content; the agent must call \texttt{page\_visit} to read a result.

\texttt{page\_visit(url: str) -> str}.  The agent submits a URL. The tool fetches the page, renders it, strips navigation and boilerplate, and returns the main textual content as a single string (typically 2{,}000--20{,}000 characters depending on the page). Images, scripts, and interactive elements are not returned.

\texttt{pdf\_fetch(url: str) -> str}.  The agent submits a URL pointing to a PDF document. The tool downloads the PDF, extracts the text layer, and returns it as a single string. Scanned-image PDFs without a text layer return an empty string.

No tool returns source-provenance information, cross-source hints, or any metadata beyond the raw page content. All three benchmark tools are available to every model configuration. For the eight non-Codex models, the tools are exposed to Claude Code CLI as benchmark Model Context Protocol (MCP) tool calls. For Codex CLI + GPT-5.5, the same benchmark tool contract is exposed through the Codex CLI host. Native search and browsing tools in both hosts are disabled.

The harness enforces a per-task budget of 200~tool calls (any mix of the three tools) and a wall-clock cap of 30~minutes. Sessions that exceed either limit are terminated, and any cell without an answer at that point receives a score of 0.

\subsection{Model Configuration}
\label{app:model_config}

All models are accessed through configured backend endpoints and paired with the same benchmark browsing tools. The full configuration table, including the backend identifier, the host mode, the temperature if exposed, the maximum output tokens, and the tool set, is reported in Table~\ref{tab:model_config}.

\begin{table}[h]
\centering
\caption{Per-model configuration used in the main experiments. The eight non-Codex models are hosted through Claude Code CLI with native browsing disabled. Codex CLI + GPT-5.5 is hosted through Codex CLI with native browsing disabled. Every model receives the benchmark-provided web search, page visit, and PDF retrieval tools. Backend identifiers are internal run identifiers and do not imply access to native search.}
\label{tab:model_config}
\small
\resizebox{\columnwidth}{!}{%
\begin{tabular}{llllll}
\toprule
\textbf{Model} & \textbf{Backend ID} & \textbf{Mode} & \textbf{Temp.} & \textbf{Max tok.} & \textbf{Tools} \\
\midrule
Codex CLI + GPT-5.5   & \texttt{codex-cli-gpt-5\_5-search}   & codex-cli & default & 32000 & benchmark tools \\
Claude Opus 4.7       & \texttt{claude-opus-4-7}              & claude-code & default & 32000 & benchmark MCP tools \\
Claude Sonnet 4.6     & \texttt{claude-sonnet-4-6}            & claude-code & default & 32000 & benchmark MCP tools \\
DeepSeek V4 Pro       & \texttt{deepseek/deepseek-v4-pro}     & claude-code & default & 32000 & benchmark MCP tools \\
GLM 5.1               & \texttt{z-ai/glm-5.1}                & claude-code & default & 32000 & benchmark MCP tools \\
DeepSeek V4 Flash     & \texttt{deepseek/deepseek-v4-flash}   & claude-code & default & 32000 & benchmark MCP tools \\
Kimi K2.6             & \texttt{openrouter-kimi-k2\_6}        & claude-code & default & 32000 & benchmark MCP tools \\
Qwen 3.6 Plus         & \texttt{qwen/qwen3.6-plus}            & claude-code & default & 32000 & benchmark MCP tools \\
MiniMax M2.7          & \texttt{minimax/minimax-m2.7}         & claude-code & default & 32000 & benchmark MCP tools \\
\bottomrule
\end{tabular}%
}
\end{table}

\subsection{Scoring Alternatives Considered}
\label{app:scoring_alternatives}

Two alternative scoring designs were considered and rejected.

A strong LLM judge reading an agent's long-form report and assigning a single task-level score would be inexpensive to run and straightforward to extend. We rejected it because prior work on LLM-as-a-judge evaluation~\cite{DBLP:conf/nips/ZhengC00WZL0LXZ23,bavaresco2025llms} reports that judge-based aggregate scores can vary noticeably across judge models on the same responses, which we view as conflicting with the goal of auditable evaluation. The per-cell rule-based scoring used in \MYBENCH~ trades flexibility for reproducibility and is intended to make human cross-checking tractable when disputes arise.

A binary task-level pass/fail would collapse the partial information that per-cell scoring exposes. Under this alternative, most or all evaluated agents would fail every task at the present difficulty, and the capability-axis analysis would not be feasible. We therefore keep the cell as the atomic scoring unit and report both aggregate and capability-sliced scores.

\section{Additional Diagnostics}
\label{app:full_results}

\subsection{Source-Provenance Distribution}

Source-provenance labels serve as an audit record during task construction. Across the 100~released tasks, 61\% of supporting source records are T1 primary sources, 24\% are T2 methodology-published research or formal industry and statistical datasets, and 15\% are T3 or lower. T3 and T4 sources are retained when they provide useful context, but cell-level reference answers prioritize T1 and T2 evidence whenever such evidence is publicly available.

\subsection{Cross-Source Agreement Distribution}

Cross-source agreement labels (\emph{consistent}, \emph{divergent}, \emph{single}) are assigned during construction to record whether a cell has independent corroboration, visible disagreement, or only one public source. The construction process records cross-source verification whenever independent public sources are available. The \emph{divergent} cells are concentrated in Calibration-type dimensions, where detecting source disagreement is part of the intended skill. The per-label distribution across all cells is reported in the full benchmark metadata release.

\subsection{Failure-Mode Counts}

Per-model failure-mode breakdowns require human annotation of a stratified sample of failing cells. The aggregate breakdown across the top-four models and the other five models is reported in Table~\ref{tab:failure_modes}; per-model counts are included in the full evaluation data.

\subsection{Budget Consumption}

Table~\ref{tab:budget} reports per-model budget consumption for scored runs. The average number of tool calls per task ranges from 24.8 (Claude Opus 4.7) to 41.5 (Codex CLI + GPT-5.5), and the average answer length ranges from 12{,}685 characters (Codex CLI + GPT-5.5) to 32{,}948 characters (MiniMax M2.7). These usage statistics are diagnostic only; the benchmark score is computed from the submitted cell answers under the rubric, with missing runs excluded from the scored set.

\begin{table}[h]
\centering
\caption{Budget consumption per model.}
\label{tab:budget}
\small
\begin{tabular}{lrr}
\toprule
\textbf{Model} & \textbf{Avg tool calls} & \textbf{Avg chars} \\
\midrule
Codex CLI + GPT-5.5   & 41.5 & 12{,}685 \\
Claude Opus 4.7       & 24.8 & 15{,}869 \\
DeepSeek V4 Pro       & 40.4 & 25{,}841 \\
GLM 5.1               & 33.3 & 20{,}316 \\
Claude Sonnet 4.6     & 27.3 & 25{,}973 \\
DeepSeek V4 Flash     & 39.8 & 20{,}951 \\
Qwen 3.6 Plus         & 37.2 & 18{,}526 \\
MiniMax M2.7          & 40.6 & 32{,}948 \\
Kimi K2.6             & 27.3 & 16{,}904 \\
\bottomrule
\end{tabular}
\end{table}

\section{Extended Related Work}
\label{app:extended_rw}

\subsection{Agentic Evaluation Beyond Web QA}

Recent agent benchmarks broaden evaluation from question answering to interactive tool use, operating-system control, and software engineering. AgentBench~\cite{DBLP:conf/iclr/0036YZXLL0DMYZ024}, ToolSandbox~\cite{DBLP:conf/naacl/LuHZANBMMLYWP25}, BFCL~\cite{DBLP:conf/icml/PatilMYJSSG25}, $\tau$-bench~\cite{yao2024taubenchbenchmarktoolagentuserinteraction}, AppWorld~\cite{DBLP:conf/acl/TrivediKHMDLGSB24}, OSWorld~\cite{DBLP:conf/nips/XieZCLZCHCSLLXZ24}, and WildBench~\cite{DBLP:conf/iclr/LinDCRPD0025} test agents in environments where success depends on stateful interaction rather than a single answer. Coding-agent benchmarks including HumanEval~\cite{chen2021evaluatinglargelanguagemodels}, SWE-bench~\cite{DBLP:conf/iclr/JimenezYWYPPN24}, LiveCodeBench~\cite{DBLP:conf/iclr/JainHGLYZWSSS25}, BigCodeBench~\cite{DBLP:conf/iclr/ZhuoVCH0WYZHPB025}, and Terminal-Bench~\cite{merrill2026terminalbenchbenchmarkingagentshard} establish that command-line harnesses are a major substrate for agent evaluation. \MYBENCH~ is complementary: it uses the agent setting to stress evidence management and derivation rather than code editing or tool invocation.

\subsection{Benchmark Construction and Judge-Based Evaluation}

AutoBencher~\cite{li2025autobencher} and BenchBuilder~\cite{li2025benchbuilder} automate benchmark construction, while MT-Bench/Chatbot Arena~\cite{DBLP:conf/nips/ZhengC00WZL0LXZ23} and large-scale studies of LLM judges~\cite{bavaresco2025llms} examine the reliability of model-based evaluation. Retrieval-augmented fact verification work such as RAFTS~\cite{yue2024rafts} is also relevant. \MYBENCH~ avoids an aggregate free-form judge as the primary metric; each answer is scored at the cell level against a stored reference value and explicit rubric.

\section{Extended Analysis}
\label{app:extended_analysis}

Additional analysis figures are reported below.

\begin{figure}[h]
\centering
\includegraphics[width=\columnwidth]{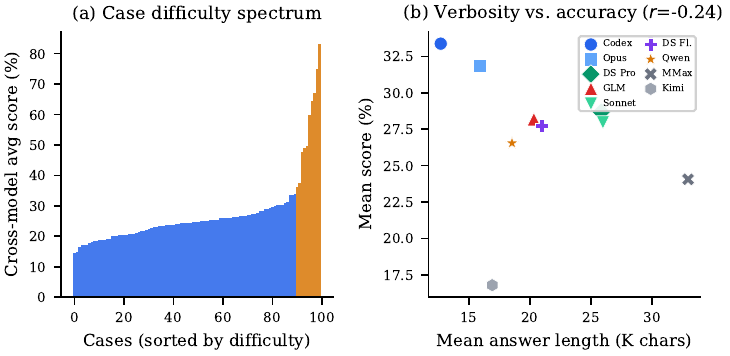}
\caption{(a) Cross-model average score distribution across 100 cases, ranked by difficulty. (b) Verbosity vs.\ accuracy ($r = -0.24$): Codex CLI + GPT-5.5 produces the shortest answers (12{,}685 characters on average) but scores highest (33.37\%); MiniMax generates 2.6$\times$ more characters for 9.3~fewer points.}
\label{fig:analysis2}
\end{figure}

\begin{figure}[h]
\centering
\includegraphics[width=\columnwidth]{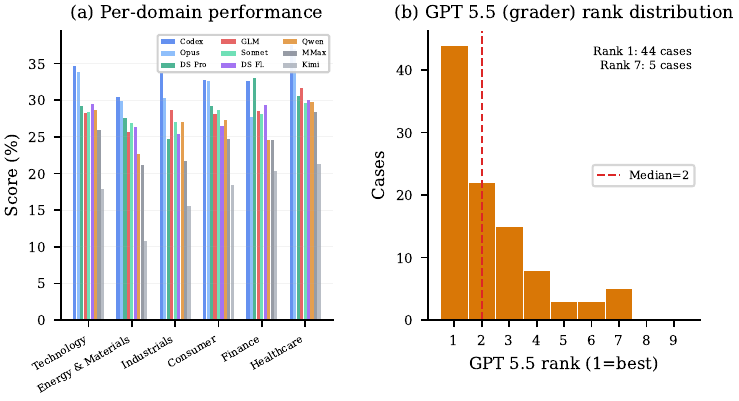}
\caption{(a) Per-domain performance across six industry categories. (b) Rank distribution of the evaluated Codex CLI + GPT~5.5 configuration across 100 cases. Although the independent grader uses the same GPT-5.5 model family, the evaluated Codex configuration does not uniformly rank first. The per-case Spearman correlation between the evaluated GPT-5.5 configuration's scores and the mean of the other eight models is $\rho = 0.73$ ($r = 0.92$ Pearson).}
\label{fig:analysis3}
\end{figure}

\begin{figure}[h]
\centering
\includegraphics[width=\columnwidth]{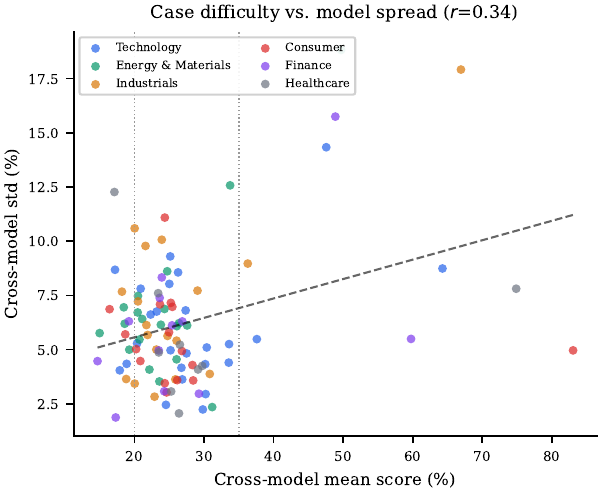}
\caption{Cross-model mean score vs.\ cross-model standard deviation per case ($r = 0.34$). Cases below the 20\% pass-band floor show compressed spread, while the 20--35\% band contains most of the benchmark and still exposes substantial model-to-model variation.}
\label{fig:spread}
\end{figure}

\begin{figure}[h]
\centering
\includegraphics[width=\columnwidth]{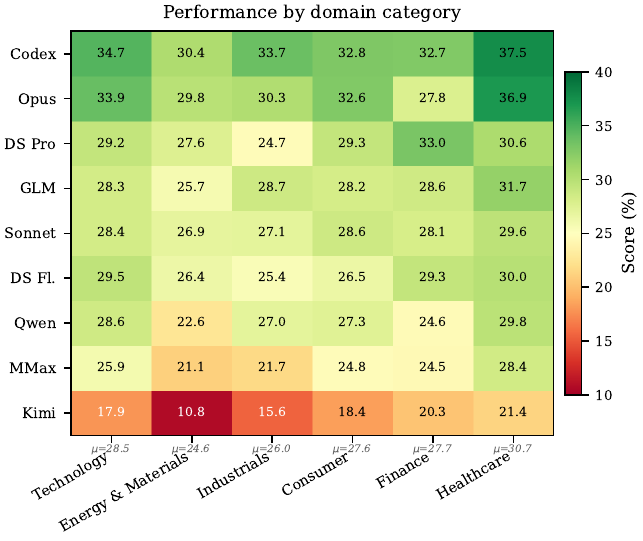}
\caption{Per-domain performance heatmap. Energy \& Materials is the hardest domain (cross-model mean 24.6\%) and Healthcare the easiest (30.7\%). Models exhibit domain-specific strengths: Codex CLI + GPT-5.5 is strongest in Healthcare and Technology, while Kimi is notably weak in Energy \& Materials.}
\label{fig:domain}
\end{figure}

\begin{figure}[h]
\centering
\includegraphics[width=\columnwidth]{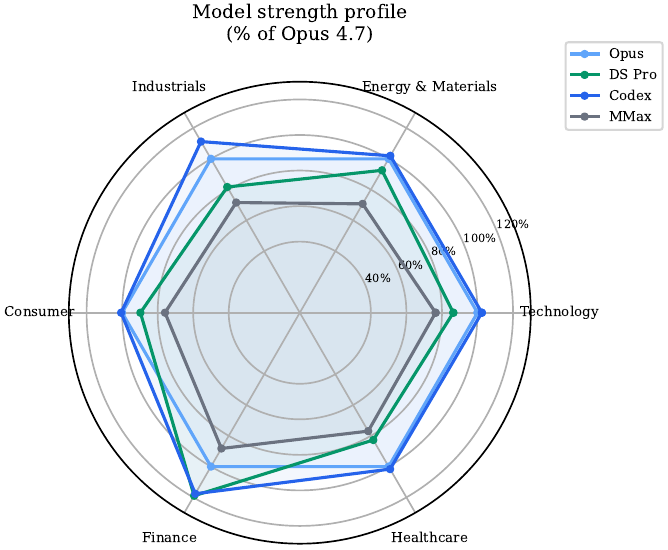}
\caption{Model strength profiles normalized to Opus~4.7's per-domain scores. Codex CLI + GPT-5.5 is relatively stronger in Finance and Industrials, DeepSeek V4 Pro is comparatively strongest in Finance, and MiniMax~M2.7 remains below Opus across all domains.}
\label{fig:radar}
\end{figure}

\begin{figure}[h]
\centering
\includegraphics[width=\columnwidth]{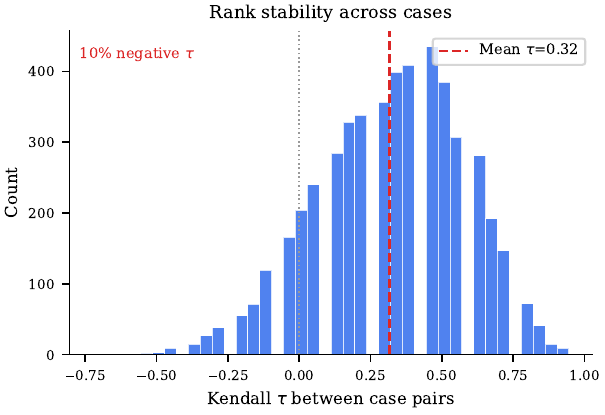}
\caption{Distribution of pairwise Kendall $\tau$ across all 4{,}950 case pairs. Mean $\tau = 0.32$ with 10\% negative values, indicating that model rankings vary across cases rather than following a fixed ordering.}
\label{fig:kendall}
\end{figure}

\begin{figure}[h]
\centering
\includegraphics[width=\columnwidth]{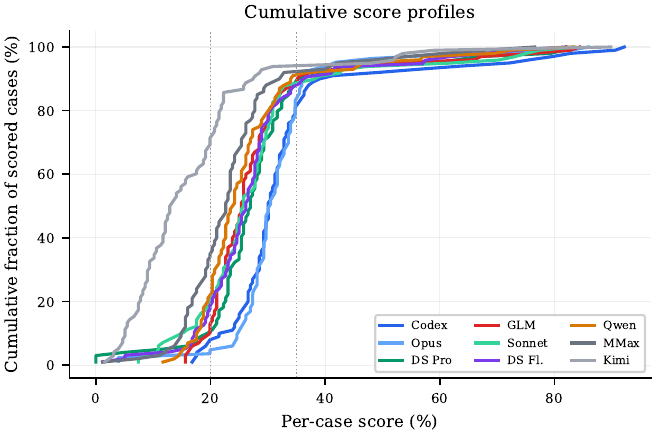}
\caption{Cumulative score profiles. Codex CLI + GPT-5.5 and Opus show the most right-shifted curves, while Kimi is left-shifted with many low-scoring cases. Qwen and MiniMax occupy the middle and lower-middle of the distribution, respectively.}
\label{fig:cdf}
\end{figure}

\section{Ethical Considerations}
\label{app:ethics}

\paragraph{Data sources and copyright.}
All reference answers in \MYBENCH~ are derived from publicly available documents: regulatory filings (SEC 10-K/20-F, Hong Kong Stock Exchange announcements), official press releases, and published research reports. The benchmark stores URLs and short factual excerpts for scoring purposes; it does not redistribute full documents. We respect robots.txt policies and do not bypass paywalls; when a publisher offers paywalled and public versions of a document, the benchmark uses only the publicly accessible material.

\paragraph{Source-provenance bias.}
The T1--T4 source-provenance scheme favors jurisdictions with mandatory public disclosure requirements (e.g., U.S. SEC, EU regulations). This may systematically disadvantage entities in markets with weaker disclosure norms. We acknowledge this limitation and encourage future extensions to incorporate regional equivalents of primary filings (e.g., China CSRC filings, India SEBI disclosures) at the T1 level.

\paragraph{Privacy.}
The benchmark contains no personally identifiable information (PII). All entities are publicly traded companies or publicly known organizations. Reference answers consist of aggregate financial and operational metrics, not individual-level data.

\paragraph{Potential positive impacts.}
\MYBENCH~ can help researchers and practitioners diagnose where deep research agents actually fail, especially on multi-step derivation, calibration, and source reconciliation rather than retrieval alone. A more discriminative benchmark may therefore support the development of more reliable research assistants for high-stakes analytical settings such as finance, policy, and scientific synthesis.

\paragraph{Potential negative impacts.}
Like other strong evaluation sets, \MYBENCH~ could be overfit to if its references are used directly for training, reducing its long-term value as an independent benchmark. More capable deep research agents can also be misused for scalable market intelligence, persuasive evidence cherry-picking, or overconfident analytical reports; this risk is one reason we emphasize auditable source records, calibrated abstention, and intended use strictly for evaluation rather than operational decision-making.

\paragraph{Intended use.}
\MYBENCH~ is intended solely for evaluating deep research agents. It should not be used to train models on the reference answers (which would constitute benchmark contamination) or to make actual investment or business decisions based on the reference values.

\section{Limitations}
\label{app:limitations}

\MYBENCH~ focuses on quantitative deep research over public documents and fixed $8 \times 8$ task matrices. It therefore does not yet cover workflows that depend on private data, non-quantitative synthesis, or extended interactive clarification with a user.

%% file: reference.bib
@article{chen2021evaluatinglargelanguagemodels,
  publtype={informal},
  author={Mark Chen and Jerry Tworek and Heewoo Jun and Qiming Yuan and Henrique Ponde de Oliveira Pinto and Jared Kaplan and Harrison Edwards and Yuri Burda and Nicholas Joseph and Greg Brockman and Alex Ray and Raul Puri and Gretchen Krueger and Michael Petrov and Heidy Khlaaf and Girish Sastry and Pamela Mishkin and Brooke Chan and Scott Gray and Nick Ryder and Mikhail Pavlov and Alethea Power and Lukasz Kaiser and Mohammad Bavarian and Clemens Winter and Philippe Tillet and Felipe Petroski Such and Dave Cummings and Matthias Plappert and Fotios Chantzis and Elizabeth Barnes and Ariel Herbert-Voss and William Hebgen Guss and Alex Nichol and Alex Paino and Nikolas Tezak and Jie Tang and Igor Babuschkin and Suchir Balaji and Shantanu Jain and William Saunders and Christopher Hesse and Andrew N. Carr and Jan Leike and Joshua Achiam and Vedant Misra and Evan Morikawa and Alec Radford and Matthew Knight and Miles Brundage and Mira Murati and Katie Mayer and Peter Welinder and Bob McGrew and Dario Amodei and Sam McCandlish and Ilya Sutskever and Wojciech Zaremba},
  title={Evaluating Large Language Models Trained on Code},
  year={2021},
  cdate={1609459200000},
  journal={CoRR},
  volume={abs/2107.03374},
  url={https://arxiv.org/abs/2107.03374}
}

@article{coelho2025deepresearchgym,
  title={Deepresearchgym: A free, transparent, and reproducible evaluation sandbox for deep research},
  author={Coelho, Jo{\~a}o and Ning, Jingjie and He, Jingyuan and Mao, Kangrui and Paladugu, Abhijay and Setlur, Pranav and Jin, Jiahe and Callan, Jamie and Magalh{\~a}es, Jo{\~a}o and Martins, Bruno and others},
  journal={arXiv preprint arXiv:2505.19253},
  year={2025}
}

@article{li2026deepresearch,
  title={DeepResearch Bench II: Diagnosing Deep Research Agents via Rubrics from Expert Report},
  author={Li, Ruizhe and Du, Mingxuan and Xu, Benfeng and Zhu, Chiwei and Wang, Xiaorui and Mao, Zhendong},
  journal={arXiv preprint arXiv:2601.08536},
  year={2026}
}

@inproceedings{DBLP:conf/iclr/JainHGLYZWSSS25,
  author       = {Naman Jain and
                  King Han and
                  Alex Gu and
                  Wen{-}Ding Li and
                  Fanjia Yan and
                  Tianjun Zhang and
                  Sida Wang and
                  Armando Solar{-}Lezama and
                  Koushik Sen and
                  Ion Stoica},
  title        = {LiveCodeBench: Holistic and Contamination Free Evaluation of Large
                  Language Models for Code},
  booktitle    = {The Thirteenth International Conference on Learning Representations,
                  {ICLR} 2025, Singapore, April 24-28, 2025},
  publisher    = {OpenReview.net},
  year         = {2025},
  url          = {https://openreview.net/forum?id=chfJJYC3iL},
  bibsource    = {dblp computer science bibliography, https://dblp.org}
}

@inproceedings{
DBLP:conf/iclr/ZhuoVCH0WYZHPB025,
title={BigCodeBench: Benchmarking Code Generation with Diverse Function Calls and Complex Instructions},
author={Terry Yue Zhuo and Vu Minh Chien and Jenny Chim and Han Hu and Wenhao Yu and Ratnadira Widyasari and Imam Nur Bani Yusuf and Haolan Zhan and Junda He and Indraneil Paul and Simon Brunner and Chen GONG and James Hoang and Armel Randy Zebaze and Xiaoheng Hong and Wen-Ding Li and Jean Kaddour and Ming Xu and Zhihan Zhang and Prateek Yadav and Naman Jain and Alex Gu and Zhoujun Cheng and Jiawei Liu and Qian Liu and Zijian Wang and David Lo and Binyuan Hui and Niklas Muennighoff and Daniel Fried and Xiaoning Du and Harm de Vries and Leandro Von Werra},
booktitle={The Thirteenth International Conference on Learning Representations},
year={2025},
url={https://openreview.net/forum?id=YrycTjllL0}
}

@inproceedings{DBLP:conf/iclr/JimenezYWYPPN24,
  author       = {Carlos E. Jimenez and
                  John Yang and
                  Alexander Wettig and
                  Shunyu Yao and
                  Kexin Pei and
                  Ofir Press and
                  Karthik R. Narasimhan},
  title        = {SWE-bench: Can Language Models Resolve Real-world Github Issues?},
  booktitle    = {The Twelfth International Conference on Learning Representations,
                  {ICLR} 2024, Vienna, Austria, May 7-11, 2024},
  publisher    = {OpenReview.net},
  year         = {2024},
  url          = {https://openreview.net/forum?id=VTF8yNQM66},
  bibsource    = {dblp computer science bibliography, https://dblp.org}
}

@inproceedings{
DBLP:conf/nips/YangJWLYNP24,
title={{SWE}-agent: Agent-Computer Interfaces Enable Automated Software Engineering},
author={John Yang and Carlos E Jimenez and Alexander Wettig and Kilian Lieret and Shunyu Yao and Karthik R Narasimhan and Ofir Press},
booktitle={The Thirty-eighth Annual Conference on Neural Information Processing Systems},
year={2024},
url={https://openreview.net/forum?id=mXpq6ut8J3}
}

@inproceedings{DBLP:conf/iclr/0001LSXTZPSLSTL25,
  author       = {Xingyao Wang and
                  Boxuan Li and
                  Yufan Song and
                  Frank F. Xu and
                  Xiangru Tang and
                  Mingchen Zhuge and
                  Jiayi Pan and
                  Yueqi Song and
                  Bowen Li and
                  Jaskirat Singh and
                  Hoang H. Tran and
                  Fuqiang Li and
                  Ren Ma and
                  Mingzhang Zheng and
                  Bill Qian and
                  Yanjun Shao and
                  Niklas Muennighoff and
                  Yizhe Zhang and
                  Binyuan Hui and
                  Junyang Lin and
                  et al.},
  title        = {OpenHands: An Open Platform for {AI} Software Developers as Generalist
                  Agents},
  booktitle    = {The Thirteenth International Conference on Learning Representations,
                  {ICLR} 2025, Singapore, April 24-28, 2025},
  publisher    = {OpenReview.net},
  year         = {2025},
  url          = {https://openreview.net/forum?id=OJd3ayDDoF},
  bibsource    = {dblp computer science bibliography, https://dblp.org}
}

@inproceedings{DBLP:conf/acl/TrivediKHMDLGSB24,
  author       = {Harsh Trivedi and
                  Tushar Khot and
                  Mareike Hartmann and
                  Ruskin Manku and
                  Vinty Dong and
                  Edward Li and
                  Shashank Gupta and
                  Ashish Sabharwal and
                  Niranjan Balasubramanian},
  editor       = {Lun{-}Wei Ku and
                  Andre Martins and
                  Vivek Srikumar},
  title        = {AppWorld: {A} Controllable World of Apps and People for Benchmarking
                  Interactive Coding Agents},
  booktitle    = {Proceedings of the 62nd Annual Meeting of the Association for Computational
                  Linguistics (Volume 1: Long Papers), {ACL} 2024, Bangkok, Thailand,
                  August 11-16, 2024},
  pages        = {16022--16076},
  publisher    = {Association for Computational Linguistics},
  year         = {2024},
  url          = {https://doi.org/10.18653/v1/2024.acl-long.850},
  doi          = {10.18653/V1/2024.ACL-LONG.850},
  bibsource    = {dblp computer science bibliography, https://dblp.org}
}

@inproceedings{DBLP:conf/iclr/0036YZXLL0DMYZ024,
  author       = {Xiao Liu and
                  Hao Yu and
                  Hanchen Zhang and
                  Yifan Xu and
                  Xuanyu Lei and
                  Hanyu Lai and
                  Yu Gu and
                  Hangliang Ding and
                  Kaiwen Men and
                  Kejuan Yang and
                  Shudan Zhang and
                  Xiang Deng and
                  Aohan Zeng and
                  Zhengxiao Du and
                  Chenhui Zhang and
                  Sheng Shen and
                  Tianjun Zhang and
                  Yu Su and
                  Huan Sun and
                  Minlie Huang and
                  Yuxiao Dong and
                  Jie Tang},
  title        = {AgentBench: Evaluating LLMs as Agents},
  booktitle    = {The Twelfth International Conference on Learning Representations,
                  {ICLR} 2024, Vienna, Austria, May 7-11, 2024},
  publisher    = {OpenReview.net},
  year         = {2024},
  url          = {https://openreview.net/forum?id=zAdUB0aCTQ},
  bibsource    = {dblp computer science bibliography, https://dblp.org}
}

@inproceedings{DBLP:conf/iclr/LinDCRPD0025,
  author       = {Bill Yuchen Lin and
                  Yuntian Deng and
                  Khyathi Raghavi Chandu and
                  Abhilasha Ravichander and
                  Valentina Pyatkin and
                  Nouha Dziri and
                  Ronan Le Bras and
                  Yejin Choi},
  title        = {WildBench: Benchmarking LLMs with Challenging Tasks from Real Users
                  in the Wild},
  booktitle    = {The Thirteenth International Conference on Learning Representations,
                  {ICLR} 2025, Singapore, April 24-28, 2025},
  publisher    = {OpenReview.net},
  year         = {2025},
  url          = {https://openreview.net/forum?id=MKEHCx25xp},
  bibsource    = {dblp computer science bibliography, https://dblp.org}
}

@inproceedings{DBLP:conf/naacl/LuHZANBMMLYWP25,
  author       = {Jiarui Lu and
                  Thomas Holleis and
                  Yizhe Zhang and
                  Bernhard Aumayer and
                  Feng Nan and
                  Haoping Bai and
                  Shuang Ma and
                  Shen Ma and
                  Mengyu Li and
                  Guoli Yin and
                  Zirui Wang and
                  Ruoming Pang},
  editor       = {Luis Chiruzzo and
                  Alan Ritter and
                  Lu Wang},
  title        = {ToolSandbox: {A} Stateful, Conversational, Interactive Evaluation
                  Benchmark for {LLM} Tool Use Capabilities},
  booktitle    = {Findings of the Association for Computational Linguistics: {NAACL}
                  2025, Albuquerque, New Mexico, USA, April 29 - May 4, 2025},
  series       = {Findings of {ACL}},
  pages        = {1160--1183},
  publisher    = {Association for Computational Linguistics},
  year         = {2025},
  url          = {https://doi.org/10.18653/v1/2025.findings-naacl.65},
  doi          = {10.18653/V1/2025.FINDINGS-NAACL.65},
  bibsource    = {dblp computer science bibliography, https://dblp.org}
}

@inproceedings{
DBLP:conf/icml/PatilMYJSSG25,
title={The Berkeley Function Calling Leaderboard ({BFCL}): From Tool Use to Agentic Evaluation of Large Language Models},
author={Shishir G Patil and Huanzhi Mao and Fanjia Yan and Charlie Cheng-Jie Ji and Vishnu Suresh and Ion Stoica and Joseph E. Gonzalez},
booktitle={Forty-second International Conference on Machine Learning},
year={2025},
url={https://openreview.net/forum?id=2GmDdhBdDk}
}

@inproceedings{DBLP:conf/nips/XieZCLZCHCSLLXZ24,
  author       = {Tianbao Xie and
                  Danyang Zhang and
                  Jixuan Chen and
                  Xiaochuan Li and
                  Siheng Zhao and
                  Ruisheng Cao and
                  Toh Jing Hua and
                  Zhoujun Cheng and
                  Dongchan Shin and
                  Fangyu Lei and
                  Yitao Liu and
                  Yiheng Xu and
                  Shuyan Zhou and
                  Silvio Savarese and
                  Caiming Xiong and
                  Victor Zhong and
                  Tao Yu},
  editor       = {Amir Globersons and
                  Lester Mackey and
                  Danielle Belgrave and
                  Angela Fan and
                  Ulrich Paquet and
                  Jakub M. Tomczak and
                  Cheng Zhang},
  title        = {OSWorld: Benchmarking Multimodal Agents for Open-Ended Tasks in Real
                  Computer Environments},
  booktitle    = {Advances in Neural Information Processing Systems 38: Annual Conference
                  on Neural Information Processing Systems 2024, NeurIPS 2024, Vancouver,
                  BC, Canada, December 10 - 15, 2024},
  year         = {2024},
  url          = {http://papers.nips.cc/paper_files/paper/2024/hash/5d413e48f84dc61244b6be550f1cd8f5-Abstract-Datasets_and_Benchmarks_Track.html},
  bibsource    = {dblp computer science bibliography, https://dblp.org}
}

@inproceedings{DBLP:conf/nips/ZhengC00WZL0LXZ23,
  author       = {Lianmin Zheng and
                  Wei{-}Lin Chiang and
                  Ying Sheng and
                  Siyuan Zhuang and
                  Zhanghao Wu and
                  Yonghao Zhuang and
                  Zi Lin and
                  Zhuohan Li and
                  Dacheng Li and
                  Eric P. Xing and
                  Hao Zhang and
                  Joseph E. Gonzalez and
                  Ion Stoica},
  editor       = {Alice Oh and
                  Tristan Naumann and
                  Amir Globerson and
                  Kate Saenko and
                  Moritz Hardt and
                  Sergey Levine},
  title        = {Judging LLM-as-a-Judge with MT-Bench and Chatbot Arena},
  booktitle    = {Advances in Neural Information Processing Systems 36: Annual Conference
                  on Neural Information Processing Systems 2023, NeurIPS 2023, New Orleans,
                  LA, USA, December 10 - 16, 2023},
  year         = {2023},
  url          = {http://papers.nips.cc/paper_files/paper/2023/hash/91f18a1287b398d378ef22505bf41832-Abstract-Datasets_and_Benchmarks.html},
  bibsource    = {dblp computer science bibliography, https://dblp.org}
}

@misc{yao2024taubenchbenchmarktoolagentuserinteraction,
      title={$\tau$-bench: A Benchmark for Tool-Agent-User Interaction in Real-World Domains},
      author={Shunyu Yao and Noah Shinn and Pedram Razavi and Karthik Narasimhan},
      year={2024},
      eprint={2406.12045},
      archivePrefix={arXiv},
      primaryClass={cs.AI},
      url={https://arxiv.org/abs/2406.12045},
}

@misc{merrill2026terminalbenchbenchmarkingagentshard,
      title={Terminal-Bench: Benchmarking Agents on Hard, Realistic Tasks in Command Line Interfaces},
      author={Mike A. Merrill and Alexander G. Shaw and Nicholas Carlini and Boxuan Li and Harsh Raj and Ivan Bercovich and Lin Shi and Jeong Yeon Shin and Thomas Walshe and E. Kelly Buchanan and Junhong Shen and Guanghao Ye and Haowei Lin and Jason Poulos and Maoyu Wang and Marianna Nezhurina and Jenia Jitsev and Di Lu and Orfeas Menis Mastromichalakis and Zhiwei Xu and Zizao Chen and Yue Liu and Robert Zhang and Leon Liangyu Chen and Anurag Kashyap and Jan-Lucas Uslu and Jeffrey Li and Jianbo Wu and Minghao Yan and Song Bian and Vedang Sharma and Ke Sun and Steven Dillmann and Akshay Anand and Andrew Lanpouthakoun and Bardia Koopah and Changran Hu and Etash Guha and Gabriel H. S. Dreiman and Jiacheng Zhu and Karl Krauth and Li Zhong and Niklas Muennighoff and Robert Amanfu and Shangyin Tan and Shreyas Pimpalgaonkar and Tushar Aggarwal and Xiangning Lin and Xin Lan and Xuandong Zhao and Yiqing Liang and Yuanli Wang and Zilong Wang and Changzhi Zhou and David Heineman and Hange Liu and Harsh Trivedi and John Yang and Junhong Lin and Manish Shetty and Michael Yang and Nabil Omi and Negin Raoof and Shanda Li and Terry Yue Zhuo and Wuwei Lin and Yiwei Dai and Yuxin Wang and Wenhao Chai and Shang Zhou and Dariush Wahdany and Ziyu She and Jiaming Hu and Zhikang Dong and Yuxuan Zhu and Sasha Cui and Ahson Saiyed and Arinbjörn Kolbeinsson and Jesse Hu and Christopher Michael Rytting and Ryan Marten and Yixin Wang and Alex Dimakis and Andy Konwinski and Ludwig Schmidt},
      year={2026},
      eprint={2601.11868},
      archivePrefix={arXiv},
      primaryClass={cs.SE},
      url={https://arxiv.org/abs/2601.11868},
}

@misc{openai2025deepresearch,
      author={{OpenAI}},
      title={Introducing deep research},
      year={2025},
      url={https://openai.com/index/introducing-deep-research/},
      note={OpenAI product announcement},
}

@misc{anthropic2025research,
      author={{Anthropic}},
      title={Claude takes research to new places},
      year={2025},
      url={https://www.anthropic.com/news/research},
      note={Anthropic product announcement},
}

@misc{google2024geminideepresearch,
      author={{Google}},
      title={Gemini: Try Deep Research and Gemini 2.0 Flash Experimental},
      year={2024},
      url={https://blog.google/products-and-platforms/products/gemini/google-gemini-deep-research/},
      note={Google product announcement},
}

@misc{perplexity2025deepresearch,
      author={{Perplexity}},
      title={Introducing Perplexity Deep Research},
      year={2025},
      url={https://www.perplexity.ai/hub/blog/introducing-perplexity-deep-research},
      note={Perplexity product announcement},
}

@misc{moonshot2025kimiresearcher,
      author={{Moonshot AI}},
      title={Kimi Researcher},
      year={2025},
      url={https://kimi.moonshot.cn/},
      note={Moonshot AI product page},
}

@misc{anthropic2025claudecode,
      author={{Anthropic}},
      title={Claude Code},
      year={2025},
      url={https://docs.anthropic.com/en/docs/claude-code/overview},
      note={Anthropic command-line coding agent documentation},
}

@misc{openai2025codexcli,
      author={{OpenAI}},
      title={Codex CLI},
      year={2025},
      url={https://github.com/openai/codex},
      note={OpenAI command-line coding agent repository},
}

@misc{wei2025browsecomp,
      author={Jason Wei and Zhiqing Sun and Spencer Papay and Scott McKinney and Jeffrey Han and Isa Fulford and Hyung Won Chung and Alex Tachard Passos and William Fedus and Amelia Glaese},
      title={BrowseComp: A Simple Yet Challenging Benchmark for Browsing Agents},
      year={2025},
      eprint={2504.12516},
      archivePrefix={arXiv},
      primaryClass={cs.CL},
      doi={10.48550/arXiv.2504.12516},
      url={https://arxiv.org/abs/2504.12516},
}

@inproceedings{mialon2024gaia,
      author={Gr{\'e}goire Mialon and Cl{\'e}mentine Fourrier and Thomas Wolf and Yann LeCun and Thomas Scialom},
      title={GAIA: a benchmark for General {AI} Assistants},
      booktitle={The Twelfth International Conference on Learning Representations},
      year={2024},
      url={https://openreview.net/forum?id=fibxvahvs3},
}

@inproceedings{wu2025webwalker,
      author={Jialong Wu and Wenbiao Yin and Yong Jiang and Zhenglin Wang and Zekun Xi and Runnan Fang and Linhai Zhang and Yulan He and Deyu Zhou and Pengjun Xie and Fei Huang},
      title={WebWalker: Benchmarking {LLM}s in Web Traversal},
      booktitle={Proceedings of the 63rd Annual Meeting of the Association for Computational Linguistics (Volume 1: Long Papers)},
      pages={10290--10305},
      publisher={Association for Computational Linguistics},
      year={2025},
      doi={10.18653/v1/2025.acl-long.508},
      url={https://aclanthology.org/2025.acl-long.508/},
}

@inproceedings{chen2023mind2web,
      author={Shijie Chen and Xiang Deng and Yu Gu and Sam Stevens and Yu Su and Huan Sun and Boshi Wang and Boyuan Zheng},
      title={Mind2Web: Towards a Generalist Agent for the Web},
      booktitle={Advances in Neural Information Processing Systems},
      year={2023},
      url={https://proceedings.neurips.cc/paper_files/paper/2023/hash/5950bf290a1570ea401bf98882128160-Abstract-Datasets_and_Benchmarks.html},
}

@misc{wei2024simpleqa,
      author={Jason Wei and Nguyen Karina and Hyung Won Chung and Yunxin Joy Jiao and Spencer Papay and Amelia Glaese and John Schulman and William Fedus},
      title={Measuring short-form factuality in large language models},
      year={2024},
      eprint={2411.04368},
      archivePrefix={arXiv},
      primaryClass={cs.CL},
      url={https://arxiv.org/abs/2411.04368},
}

@inproceedings{chen2025mlrbench,
      author={Hui Chen and Miao Xiong and Yujie Lu and Wei Han and Ailin Deng and Yufei He and Jiaying Wu and Yibo Li and Yue Liu and Bryan Hooi},
      title={{MLR-Bench: Evaluating AI Agents on Open-Ended Machine Learning Research}},
      booktitle={Advances in Neural Information Processing Systems},
      year={2025},
      url={https://mlanthology.org/neurips/2025/chen2025neurips-mlrbench/},
}

@inproceedings{zhang2024benchmarkingdata,
      author={Yuge Zhang and Qiyang Jiang and Xingyu Han and Nan Chen and Yuqing Yang and Kan Ren},
      title={Benchmarking Data Science Agents},
      booktitle={Proceedings of the 62nd Annual Meeting of the Association for Computational Linguistics (Volume 1: Long Papers)},
      pages={5677--5700},
      publisher={Association for Computational Linguistics},
      year={2024},
      doi={10.18653/v1/2024.acl-long.308},
      url={https://aclanthology.org/2024.acl-long.308/},
}

@inproceedings{li2025investorbench,
      author={Haohang Li and Yupeng Cao and Yangyang Yu and Shashidhar Reddy Javaji and Zhiyang Deng and Yueru He and Yuechen Jiang and Zining Zhu and K. P. Subbalakshmi and Jimin Huang and Lingfei Qian and Xueqing Peng and Jordan W. Suchow and Qianqian Xie},
      title={{INVESTORBENCH: A Benchmark for Financial Decision-Making Tasks with LLM-based Agent}},
      booktitle={Proceedings of the 63rd Annual Meeting of the Association for Computational Linguistics (Volume 1: Long Papers)},
      pages={2509--2525},
      publisher={Association for Computational Linguistics},
      year={2025},
      url={https://aclanthology.org/2025.acl-long.126/},
}

@article{asai2026openscholar,
  title={Synthesizing scientific literature with retrieval-augmented language models},
  author={Asai, Akari and He, Jacqueline and Shao, Rulin and Shi, Weijia and Singh, Amanpreet and Chang, Joseph Chee and Lo, Kyle and Soldaini, Luca and Feldman, Sergey and D’Arcy, Mike and others},
  journal={Nature},
  pages={1--7},
  year={2026},
  publisher={Nature Publishing Group UK London}
}

@article{phan2026humanity,
  title={A benchmark of expert-level academic questions to assess AI capabilities},
  author={Center for AI Safety Phan Long agibenchmark@ safe. ai 1 Gatti Alice 1 Li Nathaniel 1 Khoja Adam 1 Kim Ryan 1 Ren Richard 1 Hausenloy Jason 1 Zhang Oliver 1 Mazeika Mantas 1 Hendrycks Dan dan@ safe. ai 1},
  journal={Nature},
  volume={649},
  number={8099},
  pages={1139--1146},
  year={2026},
  publisher={Nature Publishing Group UK London}
}

@inproceedings{
rein2023gpqa,
title={{GPQA}: A Graduate-Level Google-Proof Q\&A Benchmark},
author={David Rein and Betty Li Hou and Asa Cooper Stickland and Jackson Petty and Richard Yuanzhe Pang and Julien Dirani and Julian Michael and Samuel R. Bowman},
booktitle={First Conference on Language Modeling},
year={2024},
url={https://openreview.net/forum?id=Ti67584b98}
}

@inproceedings{li2025autobencher,
      author={Xiang Lisa Li and Farzaan Kaiyom and Evan Zheran Liu and Yifan Mai and Percy Liang and Tatsunori Hashimoto},
      title={AutoBencher: Towards Declarative Benchmark Construction},
      booktitle={The Thirteenth International Conference on Learning Representations},
      year={2025},
      url={https://openreview.net/forum?id=ymt4crbbXh},
}

@inproceedings{li2025benchbuilder,
      author={Tianle Li and Wei-Lin Chiang and Evan Frick and Lisa Dunlap and Tianhao Wu and Banghua Zhu and Joseph E. Gonzalez and Ion Stoica},
      title={From Crowdsourced Data to High-quality Benchmarks: Arena-Hard and BenchBuilder Pipeline},
      booktitle={Forty-second International Conference on Machine Learning},
      year={2025},
      url={https://openreview.net/forum?id=KfTf9vFvSn},
}

@inproceedings{bavaresco2025llms,
      author={Anna Bavaresco and Raffaella Bernardi and Leonardo Bertolazzi and Desmond Elliott and Raquel Fern{\'a}ndez and Albert Gatt and Esam Ghaleb and Mario Giulianelli and Michael Hanna and Alexander Koller and Andr{\'e} F. T. Martins and Philipp Mondorf and Vera Neplenbroek and Sandro Pezzelle and Barbara Plank and David Schlangen and Alessandro Suglia and Aditya K. Surikuchi and Ece Takmaz and Alberto Testoni},
      title={{LLMs instead of Human Judges? A Large Scale Empirical Study across 20 NLP Evaluation Tasks}},
      booktitle={Proceedings of the 63rd Annual Meeting of the Association for Computational Linguistics (Volume 2: Short Papers)},
      pages={238--255},
      publisher={Association for Computational Linguistics},
      year={2025},
      url={https://aclanthology.org/2025.acl-short.20/},
}

@inproceedings{yue2024rafts,
      author={Zhenrui Yue and Huimin Zeng and Lanyu Shang and Yifan Liu and Yang Zhang and Dong Wang},
      title={Retrieval Augmented Fact Verification by Synthesizing Contrastive Arguments},
      booktitle={Proceedings of the 62nd Annual Meeting of the Association for Computational Linguistics (Volume 1: Long Papers)},
      publisher={Association for Computational Linguistics},
      year={2024},
      doi={10.18653/v1/2024.acl-long.556},
      url={https://aclanthology.org/2024.acl-long.556/},
}

@misc{du2025deepresearchbench,
      title={DeepResearch Bench: A Comprehensive Benchmark for Deep Research Agents},
      author={Mingxuan Du and Benfeng Xu and Chiwei Zhu and Xiaorui Wang and Zhendong Mao},
      year={2025},
      eprint={2506.11763},
      archivePrefix={arXiv},
      primaryClass={cs.CL},
      url={https://arxiv.org/abs/2506.11763},
}

@misc{gupta2026deepsearchqa,
      title={DeepSearchQA: Bridging the Comprehensiveness Gap for Deep Research Agents},
      author={Nikita Gupta and Riju Chatterjee and Lukas Haas and Connie Tao and Andrew Wang and Chang Liu and Hidekazu Oiwa and Elena Gribovskaya and Jan Ackermann and John Blitzer and Sasha Goldshtein and Dipanjan Das},
      year={2026},
      eprint={2601.20975},
      archivePrefix={arXiv},
      primaryClass={cs.IR},
      url={https://arxiv.org/abs/2601.20975},
}

@misc{zhong2026draco,
      title={DRACO: a Cross-Domain Benchmark for Deep Research Accuracy, Completeness, and Objectivity},
      author={Joey Zhong and Hao Zhang and Clare Southern and Jeremy Yang and Thomas Wang and Kate Jung and Shu Zhang and Denis Yarats and Johnny Ho and Jerry Ma},
      year={2026},
      eprint={2602.11685},
      archivePrefix={arXiv},
      primaryClass={cs.CL},
      url={https://arxiv.org/abs/2602.11685},
}

@misc{wong2025widesearch,
      title={WideSearch: Benchmarking Agentic Broad Info-Seeking},
      author={Ryan Wong and Jiawei Wang and Junjie Zhao and Li Chen and Yan Gao and Long Zhang and Xuan Zhou and Zuo Wang and Kai Xiang and Ge Zhang and Wenhao Huang and Yang Wang and Ke Wang},
      year={2025},
      eprint={2508.07999},
      archivePrefix={arXiv},
      primaryClass={cs.IR},
      url={https://arxiv.org/abs/2508.07999},
}

@misc{abaskohi2025drbench,
      title={DRBench: A Realistic Benchmark for Enterprise Deep Research},
      author={Amirhossein Abaskohi and Tianyi Chen and Miguel Mu{\~n}oz-M{\'a}rmol and Curtis Fox and Amrutha Varshini Ramesh and {\'E}tienne Marcotte and Xing Han L{\`u} and Nicolas Chapados and Spandana Gella and Peter West and Giuseppe Carenini and Christopher Pal and Alexandre Drouin and Issam H. Laradji},
      year={2025},
      eprint={2510.00172},
      archivePrefix={arXiv},
      primaryClass={cs.CL},
      url={https://arxiv.org/abs/2510.00172},
}

@misc{wang2025liveresearchbench,
      title={LiveResearchBench: A Live Benchmark for User-Centric Deep Research in the Wild},
      author={Jiayu Wang and Yifei Ming and Riya Dulepet and Qinglin Chen and Austin Xu and Zixuan Ke and Frederic Sala and Aws Albarghouthi and Caiming Xiong and Shafiq Joty},
      year={2025},
      eprint={2510.14240},
      archivePrefix={arXiv},
      primaryClass={cs.CL},
      url={https://arxiv.org/abs/2510.14240},
}

@misc{browsecompzh2025,
      title={BrowseComp-ZH: Benchmarking Web Browsing Ability of Large Language Models in Chinese},
      author={Peilin Zhou and Bruce Leon and Xiang Ying and Can Zhang and Yifan Shao and Qichen Ye and Dading Chong and Zhiling Jin and Chenxuan Xie and Meng Cao and Yuxin Gu and Sixin Hong and Jing Ren and Jian Chen and Chao Liu and Yining Hua},
      year={2025},
      eprint={2504.19314},
      archivePrefix={arXiv},
      primaryClass={cs.CL},
      url={https://arxiv.org/abs/2504.19314},
}

@misc{anthropic2026opus47,
      author={{Anthropic}},
      title={Introducing Claude Opus 4.7},
      year={2026},
      url={https://www.anthropic.com/news/claude-opus-4-7},
      note={Anthropic product announcement, April 16, 2026},
}

@misc{anthropic2026sonnet46,
      author={{Anthropic}},
      title={Introducing Claude Sonnet 4.6},
      year={2026},
      url={https://www.anthropic.com/news/claude-sonnet-4-6},
      note={Anthropic product announcement, February 17, 2026},
}

@misc{deepseek2026v4preview,
      author={{DeepSeek AI}},
      title={DeepSeek V4 Preview Release},
      year={2026},
      url={https://api-docs.deepseek.com/news/news260424},
      note={DeepSeek API documentation news, April 24, 2026},
}

@misc{zai2026glm51,
      author={{Z.AI}},
      title={{GLM-5.1} Overview},
      year={2026},
      url={https://docs.z.ai/guides/llm/glm-5.1},
      note={Z.AI developer documentation},
}

@misc{moonshot2026kimik26,
      author={{Moonshot AI}},
      title={Kimi K2.6},
      year={2026},
      url={https://www.kimi.com/ai-models/kimi-k2-6},
      note={Moonshot AI model page},
}

@misc{openai2026gpt55,
      author={{OpenAI}},
      title={Introducing GPT-5.5},
      year={2026},
      url={https://openai.com/index/introducing-gpt-5-5/},
      note={OpenAI product release, April 23, 2026},
}

@misc{qwen2026qwen36plus,
      author={{Qwen Team}},
      title={Qwen3.6-Plus: Towards Real World Agents},
      year={2026},
      url={https://qwen.ai/blog?id=qwen3.6},
      note={Qwen product announcement, April 1, 2026},
}

@misc{minimax2026m27,
      author={{MiniMax}},
      title={MiniMax M2.7},
      year={2026},
      url={https://www.minimax.io/models/text/m27},
      note={MiniMax model page},
}

@misc{wu2026deepresearch9k,
      title={{DeepResearch-9K}: A Challenging Benchmark Dataset of Deep-Research Agent},
      author={Tongzhou Wu and others},
      year={2026},
      eprint={2603.01152},
      archivePrefix={arXiv},
      url={https://arxiv.org/abs/2603.01152},
}

@inproceedings{
gou2025mind2web2,
title={Mind2Web 2: Evaluating Agentic Search with Agent-as-a-Judge},
author={Boyu Gou and Zanming Huang and Yuting Ning and Yu Gu and Michael Lin and Weijian Qi and Andrei Kopanev and Botao Yu and Bernal Jimenez Gutierrez and Yiheng Shu and Chan Hee Song and Jiaman Wu and Shijie Chen and Hanane Nour Moussa and TIANSHU ZHANG and Jian Xie and Yifei Li and Tianci Xue and Zeyi Liao and Kai Zhang and Boyuan Zheng and Zhaowei Cai and Viktor Rozgic and Morteza Ziyadi and Huan Sun and Yu Su},
booktitle={The Thirty-ninth Annual Conference on Neural Information Processing Systems Datasets and Benchmarks Track},
year={2026},
url={https://openreview.net/forum?id=AUaW6DS9si}
}

@misc{li2026openresearcher,
      title={{OpenResearcher}: A Fully Open Pipeline for Long-Horizon Deep Research Trajectory Synthesis},
      author={Zhuofan Li and others},
      year={2026},
      eprint={2603.20278},
      archivePrefix={arXiv},
      url={https://arxiv.org/abs/2603.20278},
}

@misc{you2026autoresearchbench,
      title={{AutoResearchBench}: Benchmarking {AI} Agents on Complex Scientific Literature Discovery},
      author={Cher You and Bowen Chen and Xuan Wang and others},
      year={2026},
      eprint={2604.25256},
      archivePrefix={arXiv},
      url={https://arxiv.org/abs/2604.25256},
}
